\documentclass{article}

\usepackage{graphicx} 
\usepackage{float}
\usepackage{authblk}

\usepackage[letterpaper,top=2cm,bottom=2cm,left=3cm,right=3cm,marginparwidth=1.75cm]{geometry}
\usepackage{algorithm}
\usepackage{algpseudocode}
\usepackage{amsmath}
\usepackage{multirow}
\usepackage{doi}
\usepackage{booktabs}
\usepackage{float}
\usepackage{makecell}
\usepackage{CJKutf8}
\usepackage[numbers,sort&compress]{natbib}
\usepackage{enumitem}

\title{Opportunities and Challenges of Large Language Models for Low-Resource Languages in Humanities Research}

\newcommand{\equalcontrib}{\textsuperscript{*}}

\author[1,2]{Tianyang Zhong\thanks{Co-first author}}
\author[1]{Zhenyuan Yang\equalcontrib}
\author[1]{Zhengliang Liu\equalcontrib}
\author[1,3]{Ruidong Zhang\equalcontrib}
\author[1]{Weihang You}
\author[1]{Yiheng Liu}
\author[1]{Haiyang Sun}
\author[1]{Yi Pan}
\author[1]{Yiwei Li}
\author[1]{Yifan Zhou}
\author[1]{Hanqi Jiang}
\author[1]{Junhao Chen}
\author[1]{Peng Shu}
\author[1]{Shaochen Xu}
\author[1]{Zihao Wu}
\author[1]{Huaqin Zhao}
\author[1]{Wei Ruan}
\author[1]{Xinliang Li}

\author[4]{Xiang Li}
\author[1]{Tianming Liu\thanks{Corresponding author}}

\affil[1]{School of Computing, The University of Georgia, Athens 30602, USA}
\affil[2]{Department of Mathematical and Statistical Sciences, University of Alberta, Edmonton, Canada}
\affil[3]{University of California, Los Angeles, CA, USA}
\affil[4]{Department of Radiology, Massachusetts General Hospital and Harvard Medical School, MA, USA}

\begin{document}
\date{}

\maketitle
\begin{abstract}
Low-resource languages serve as invaluable repositories of human history, embodying cultural evolution and intellectual diversity. Despite their significance, these languages face critical challenges, including data scarcity and technological limitations, which hinder their comprehensive study and preservation. Recent advancements in large language models (LLMs) offer transformative opportunities for addressing these challenges, enabling innovative methodologies in linguistic, historical, and cultural research. This study systematically evaluates the applications of LLMs in low-resource language research, encompassing linguistic variation, historical documentation, cultural expressions, and literary analysis. By analyzing technical frameworks, current methodologies, and ethical considerations, this paper identifies key challenges such as data accessibility, model adaptability, and cultural sensitivity. Given the cultural, historical, and linguistic richness inherent in low-resource languages, this work emphasizes interdisciplinary collaboration and the development of customized models as promising avenues for advancing research in this domain. By underscoring the potential of integrating artificial intelligence with the humanities to preserve and study humanity's linguistic and cultural heritage, this study fosters global efforts towards safeguarding intellectual diversity.

\end{abstract}

\section{Introduction}
\subsection{Research Background}
\subsubsection{Importance and Endangerment of Low-Resource Languages in the Global Linguistic Ecology}

The linguistic landscape of the world constitutes a complex tapestry interwoven with a rich diversity of languages, each strand epitomizing a distinctive cultural, historical, and social identity. This global linguistic diversity forms a foundational pillar of human civilization, cultivating an array of perspectives and worldviews that enhance our collective intellectual legacy. Among these, low-resource languages occupy a particularly crucial position, not merely as modes of communication but as repositories of distinctive cultural knowledge, historical narratives, and worldviews. These languages, frequently spoken by smaller communities, are essential to the preservation of cultural heritage and the transmission of indigenous knowledge systems.

However, the global linguistic landscape is presently undergoing an extraordinary crisis, with low-resource languages among the most threatened. The swift vanishing of these languages is of serious concern, highlighted by concerning data and studies. It is estimated, for example, that around 40\% of the world's 7,000 languages face extinction, with numerous low-resource languages having fewer than 1,000 speakers \cite{vasconcelos2024disappearing}. This decline is caused by several factors, such as the widespread effects of globalization, the challenges of urbanization, and the prevalence of dominant languages. These elements often marginalize low-resource languages, resulting in their gradual decline and eventual disappearance.

The importance of low-resource languages, both culturally and intellectually, is immense. These languages are not simply tools for conversation; they serve as vibrant repositories of human history, reflecting centuries of cultural and intellectual development. When a low-resource language disappears, it is as though an entire library fades away, taking with it distinctive stories, oral histories, and scientific insights. Such a cultural and intellectual loss deeply impacts humanity, stripping future generations of the diverse richness of human thought and experience these languages offer. Consequently, preserving and revitalizing low-resource languages are not only academic pursuits but also critical actions needed to protect our global cultural heritage.

\subsubsection{Demand for Low-Resource Languages in Humanities Research and the Insufficiencies in Existing Studies}

Languages with limited resources are crucial for advancing research in the humanities, especially in fields like anthropology, history, literature, and linguistics. These languages provide distinctive perspectives on cultures, societies, and intellectual traditions that are less frequently studied, thus broadening our appreciation of human diversity. For example, anthropologists frequently use indigenous languages to reveal hidden facets of cultural practices and social organizations not captured by dominant languages. Likewise, historians and literary scholars gain access to primary sources and oral histories maintained in low-resource languages, offering a more detailed and thorough understanding of historical events and literary traditions. In the field of linguistics, analyzing low-resource languages aids the creation of theories and models applicable to a wider variety of human languages, thereby deepening our knowledge of language universals and diversity \cite{gwerevende2023safeguarding}.

The critical significance of low-resource languages notwithstanding, prevailing research methodologies encounter appreciable challenges when engaging with these languages. A principal limitation is the paucity of textual data. Numerous low-resource languages lack extensive written corpora, thereby complicating the execution of empirical studies and the development of computational models. Furthermore, the absence of specialized computational tools specifically adapted to these languages aggravates the issue \cite{liu2023transformation}. Existing tools and resources are frequently designed for high-resource languages, thereby proving ineffective or inefficient for application to low-resource languages. As a brief empirical note from our working list, coverage remains uneven across LLM systems. For example, GPT5 supports \emph{Manchu} whereas NLLB does not, while NLLB supports \emph{Istro-Venetian} but GPT5 is not accurate. In short, each system covers some different languages, yet overall support for low-resource, including historical and regional, varieties remains insufficient for dependable humanities workflows. This technological disparity is further heightened by the inadequate availability of essential linguistic resources such as dictionaries, grammars, and annotated corpora, which are imperative for both descriptive and computational linguistic research.

The literature concerning low-resource languages is marked by a significant disparity, with a disproportionate emphasis on high-resource languages. Such a bias constrains our comprehension of human culture and history, as it neglects the contributions of lesser-studied languages and cultures. The predominant focus on high-resource languages may lead to a skewed perspective, wherein certain cultural and historical narratives are accorded undue prominence over others. This imbalance not only compromises the integrity of humanities research but also perpetuates a form of cultural erasure. To rectify this gap, there exists an urgent requirement for the development of novel research methodologies and tools that can effectively integrate low-resource languages into scholarly discourse. Achieving this would facilitate a more inclusive and comprehensive understanding of human culture and history, ultimately allowing humanities research to realize its potential to illuminate the full spectrum of human experience.

\subsection{Opportunities for Low-Resource Language Research Through Large Language Models}

\subsubsection{Breakthroughs in Language Processing with LLMs }

The emergence of LLMs marks a new phase in natural language processing (NLP), significantly altering the field with their remarkable capabilities and adaptability. Examples of LLMs, such as GPT-4 \cite{achiam2023gpt} and LLaMA \cite{dubey2024llama}, are based on the transformer architecture. This architecture uses self-attention mechanisms to efficiently process and generate text with impressive fluency and contextual awareness. By supporting parallel processing of sequences, transformers overcome the challenges faced by earlier architectures like Recurrent Neural Networks (RNNs) \cite{cho2014learning} and Long Short-Term Memory (LSTM) networks \cite{graves2012long}, which struggled with handling long-range dependencies and parallelization.

The advancements made by LLMs across several NLP tasks are remarkable. These models have set new benchmarks in areas such as machine translation, text generation, sentiment analysis, and language comprehension. For example, the BERT model \cite{devlin2018bert}, which originates from the transformer architecture, has greatly improved the effectiveness and scalability of NLP systems through self-supervised pretraining. In recent developments, generative models like the GPT series \cite{radford2019language, brown2020language} have demonstrated their zero-shot and few-shot in-context learning capabilities, allowing them to tackle various tasks without the need for task-specific fine-tuning. This flexibility highlights the extensive datasets used for pretraining these models, which incorporate a vast array of world knowledge and emerging skills.

An important advancement in the field of LLMs is their multilingual capability. These models can handle and produce text in a variety of languages, including those with limited resources. Unlike traditional approaches that rely heavily on large amounts of parallel text for accurate translation, LLMs utilize extensive pretraining across many languages, enabling them to create coherent outputs even for languages with limited training data. This feature is particularly crucial for low-resource languages, which often have limited text corpora and lack specialized computational tools. By using the multilingual pretraining of LLMs, researchers now have access to more powerful and versatile tools for processing and analyzing low-resource languages, helping to close the gap in current research practices.

The incorporation of LLMs into numerous practical applications is capturing growing interest, significantly affecting a variety of areas including education, healthcare, and robotics \cite{liu2023summary,zhao2023brain,ma2024iterative,dai2023chataug,liu2023radiology,liao2023differentiating,liu2023context,rezayi2022clinicalradiobert,dai2023ad,zhao2023ophtha,zhang2024generalist,liu2023radonc,liu2024fine,lyu2024gp,wang2024comprehensive,huang2024position,liu2024surviving,huang2024trustllm,tian2024assessing}. Their capacity for human-like comprehension and reasoning is a stepping stone toward Artificial General Intelligence (AGI), fostering societal progress across numerous disciplines\cite{zhao2023brain,liu2023transformation,zhenyuan2024analyzing,zhao2024revolutionizing,li2024large,lee2023multimodality,wang2024legal}. As LLMs advance, they hold vast potential to enrich research in languages with limited resources, providing novel opportunities to preserve and comprehend the extensive cultural and intellectual treasures embedded in these languages.

\subsubsection{Prospects and Challenges for Applications in Low-Resource Language Research}

Integrating LLMs into research on low-resource languages presents numerous possibilities for enhancing understanding and conservation efforts associated with these languages. Models such as GPT-3, GPT-4, and LLaMA have demonstrated capabilities in areas like text generation, translation, sentiment analysis, and linguistic analysis. For instance, LLMs could generate coherent and contextually relevant narratives in low-resource languages, which might aid in documenting and sharing oral histories and cultural stories. In terms of translation, these models may help enable cross-linguistic communication by producing somewhat accurate and fluent translations, especially useful for languages with limited parallel corpora. Furthermore, sentiment and linguistic analyses could provide valuable insights into the emotional and structural nuances of these languages, supporting a more comprehensive understanding from linguistic and anthropological standpoints.

However, applying LLMs to low-resource languages presents certain challenges. A major hurdle is the scarcity of data. These languages often lack a substantial body of written material, which complicates the training and refinement of LLMs. This shortage may result in less than optimal performance, especially in languages with intricate linguistic aspects or minimal presence in the training datasets. Additionally, model bias is a notable concern. Although LLMs are pretrained on extensive and varied data collections, these might not adequately capture low-resource languages, culminating in skewed or incorrect results. Furthermore, the challenge of obtaining specialized training data persists. Although LLMs can undertake zero-shot or few-shot tasks via prompt engineering, their performance is notably improved with task-specific fine-tuning, necessitating extra data that might not be readily accessible for low-resource languages.

Current research endeavors are actively tackling these challenges through a range of techniques. Strategies like fine-tuning, transfer learning, and data augmentation are being investigated for adapting LLMs to low-resource languages. Fine-tuning entails training the model with a smaller, task-specific dataset, enhancing its performance on low-resource languages by customizing it to their unique features. Transfer learning utilizes the insights gained by LLMs during pretraining and applies them to novel, related tasks, thereby minimizing the requirement for a large amount of training data. Data augmentation approaches, including back-translation, have become prevalent in low-resource machine translation. Back-translation uses a model to convert monolingual target language data into the source language, forming pseudo-parallel corpora that substantially increase the amount of training data for low-resource pairs. Unsupervised machine translation, which completely removes the need for parallel data, has emerged as a promising solution for low-resource languages by using monolingual corpora from both the source and target languages and developing the ability to align and translate between them via iterative back-translation and shared latent representations.

Notwithstanding these advancements, the implementation of LLMs in the context of low-resource languages persists as a subject of active scholarly investigation. The evolution of LLMs has unveiled novel opportunities for translation of low-resource languages, facilitating the generation of coherent textual outputs even in languages characterized by scarcity of training data. Nevertheless, the efficacy of these models in addressing the intricacies and unique linguistic attributes inherent to low-resource languages, particularly those with minimal representation in training datasets, remains a significant challenge. As research in this domain advances, it is imperative to persist in the exploration of innovative methodologies and tools that can effectively integrate low-resource languages into academic discourse, thereby augmenting our comprehension of human culture and history.

\subsection{Research Objectives and Contributions}

\subsubsection{Investigation of Opportunities and Challenges of LLMs in Humanities Research on Low-Resource Languages}

The main goal of this study is to thoroughly examine the potential benefits and obstacles associated with using LLMs for humanities research focused on low-resource languages. This goal is motivated by the pressing necessity to preserve and investigate these languages, as they hold distinctive cultural, historical, and social insights. By investigating how LLMs can advance research in this field, we aspire to make a valuable contribution to the wider spectrum of humanities research, thereby deepening our comprehension of human diversity and cultural legacy.

\begin{figure}[H]
    \centerline{\includegraphics[width=\columnwidth,height=0.6\columnwidth]{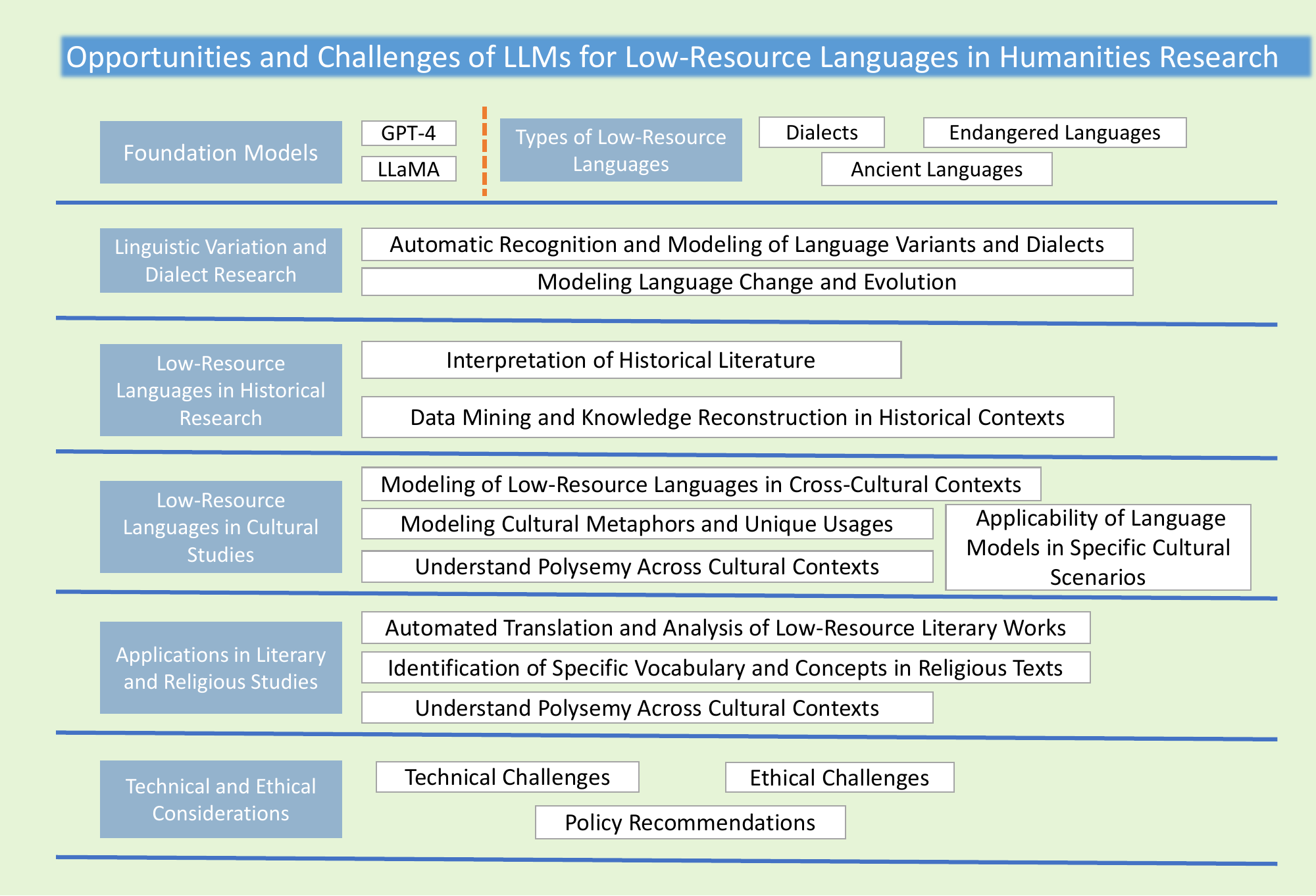}}
    \caption{Overview of the structure outline of the article.}
    \label{fig:outline}
\end{figure}

The main content of the article is summarized in the figure~\ref{fig:outline}. This research aims to deliver substantial and diverse contributions. Primarily, it will offer an in-depth review of the current state of LLMs in the context of low-resource language research. This review will provide a detailed assessment of existing methodologies, tools, and datasets, and will spotlight both progress made and existing challenges in the field. Furthermore, we aim to pinpoint specific ways in which LLMs can substantially influence research in the humanities. These applications cover areas such as text generation, translation, sentiment analysis, and linguistic examination. Moreover, we'll delve into the potential for LLMs in areas like language variation and historical, cultural, and religious studies. For example, LLMs can aid in translating and interpreting ancient texts, analyzing linguistic changes and variations, and understanding cultural and religious subtleties in low-resource languages.

This research will critically examine the hurdles and constraints linked to utilizing LLMs for low-resource languages, marking a significant contribution to the field. Among these hurdles are limited data availability, model bias, the requirement for specialized datasets, and the ethical considerations of deploying LLMs in culturally delicate situations. By tackling these issues, the study aims to provide a balanced view of the potential and limitations of LLMs in this area. Additionally, it will propose recommendations for future research and practical applications, highlighting innovative methods and tools that can effectively integrate low-resource languages into academic discussions. 

The importance of this research lies in its potential to aid in the preservation and study of low-resource languages, which serve as repositories of human history, capturing centuries of cultural evolution and intellectual progress. The disappearance of a low-resource language can be likened to losing an entire library, with its distinct stories, oral traditions, and scientific knowledge fading away. By harnessing the power of LLMs, we can improve our ability to document, study, and preserve these languages, protecting our global cultural heritage. This work is also crucial in the broader context of humanities research, as it opens up new possibilities for understanding the diverse tapestry of human thought and experience. In conclusion, this research aims to connect the advances in LLMs with the urgent necessity of preserving and studying low-resource languages. By exploring both the opportunities and hurdles presented by LLMs, we hope to advance the creation of innovative methods and tools that can effectively incorporate low-resource languages into academic discourse. This will, in turn, expand our comprehension of human culture and history, meeting the humanities' promise to shed light on the full range of human experience.

\section{Linguistic Variation and Dialect Research in Low-Resource Languages}

In the field of sociolinguistics, research on under-described languages has traditionally been constrained to specific tasks such as part-of-speech tagging, text classification, and machine translation\cite{magueresse2020lowresourcelanguagesreviewpast}. However, the emergence of LLMs, as outlined in the foundational framework, has transformed this landscape. By leveraging their computational power, reasoning capabilities, and advanced methodologies, LLMs confront the challenges posed by limited documented linguistic resources, enabling more comprehensive research on linguistic variation and dialects in low-resource languages.

Unlike widely used corpus-based approaches in linguistics, which struggle to handle low-resource languages due to data scarcity, LLMs-based methods utilize advanced techniques to effectively process diverse language variants. Through these innovations, LLMs play a crucial role in bridging the gaps left by conventional methods, opening new possibilities for understanding and preserving linguistic diversity in under-described languages. These innovations open new pathways for understanding and preserving the linguistic diversity of under-described languages while addressing critical gaps in research methodologies.

\subsection{Automatic Recognition and Modeling of Language Variants and Dialects}

Recognizing and modeling language variants and dialects in low-resource contexts is a complex but crucial task in NLP.  LLMs address these challenges by leveraging multilingual datasets and employing techniques such as prompt engineering, retrieved-augmented generation (RAG), and meta-learning that allow the linguistic system to adapt to specific dialects and generalize across diverse linguistic contexts, even with small training data size.

\subsubsection{Existing Methods of Handling Diverse Variants in LLMs}
Several structured methods have been developed to enhance LLMs pipelines for processing various language variants and dialects. These approaches focus on building efficient systems for identifying, modeling, and leveraging linguistic diversity:

\textbf{Transfer Learning and Fine-Tuning}
LLMs utilize transfer learning to adapt knowledge from high-resource languages to low-resource languages\cite{alam2024llms}, effectively bridging linguistic gaps. Fine-tuning on small, domain-specific datasets helps capture regional and dialectal nuances. Techniques such as Low-Rank Adaptation (LoRA) and QLoRA further enhance fine-tuning efficiency, making this approach computationally accessible for low-resource settings.

\textbf{Prompt Engineering and In-Context Learning}
Prompt engineering is crucial for tailoring LLMs outputs to specific dialectal contexts. Minor modifications to the prompts, such as the adjustment of word order or structure, can significantly influence the model output\cite{parida2024pretrain}. In-context learning further supports low-resource scenarios by enabling LLMs to adapt without requiring extensive retraining, leveraging contextual examples to generate more accurate responses.

\textbf{Retrieval-Augmented Generation}
RAG enhances the performance of LLMs by integrating retrieval mechanisms that fetch relevant contextual data, which is especially valuable for underrepresented languages. It is particularly valuable for underrepresented languages, where contextual augmentation can compensate for limited data availability\cite{seo2024retrievalaugmenteddataaugmentationlowresource}. In addition to leveraging retrieval data to enhance training datasets, there are advanced RAG approaches that dynamically select the appropriate dialect context from external knowledge bases or corpora to enhance generation accuracy.

\textbf{Meta-Learning}
Meta-learning improves the adaptability of LLMs, allowing them to generalize across multiple tasks and linguistic variations. By training on a variety of tasks, meta-learning empowers LLMs to recognize unseen dialects and adapt to new linguistic contexts with minimal labeled data. This capability is invaluable for under-described languages, where conventional training often fails to achieve desired accuracy\cite{10288436}.

\subsubsection{Dialect Speech Recognition in Low-Resource Contexts}
Speech recognition systems are indispensable for addressing dialect recognition in low-resource settings, where annotated datasets are scarce. Advanced methods such as the state-of-the-art Conformer Transducer (ConfT) model, combined with Model-Agnostic Meta-Learning (MAML), have demonstrated substantial improvements in dialect-specific tasks. These methods optimize parameters for rapid fine-tuning with minimal data, reducing word error rates by up to 12\% for low-resource accents, as evidenced in recent studies\cite{romanenko2020robust} \cite{10485786}. Building on advancements in speech-to-text and text-to-speech models, LLMs can further enhance dialect recognition by integrating phonetic and acoustic features into language understanding pipelines. By leveraging dialect-specific lexicons and fine-tuned language models, these systems achieve improved transcription accuracy, making them vital tools for the preservation and accessibility of linguistic diversity.

\subsubsection{Challenges in Dialect Recognition and Modeling}
Despite technological developments, significant challenges persist in the recognition and modeling of dialects, particularly in low-resource contexts based on dialect processing:

\begin{itemize}
    \item \textbf{Annotation Gaps}: Limited or poorly annotated datasets restrict the model's performance and generalization to diverse linguistic contexts.
\item \textbf{Inconsistencies in Spoken and Written Forms}: Variations in grammar, orthography, and style complicate model training and evaluation, often leading to misinterpretations and reduced accuracy.
\item \textbf{Biases and Sampling Limitations}: Over-representation\cite{terhoeve2022highresourcemethodologicalbiaslowresource} of high-resource languages in training data skews model output, undermining efforts to build inclusive systems capable of capturing the full spectrum of linguistic variation.
\item \textbf{Complex Acoustic Features in Dialects}: The unique phonological and prosodic features require specialized modeling techniques for dialects. Speech recognition systems must account for these variations, which often necessitate additional training resources and strategies.
\end{itemize}

Addressing these challenges requires a multifaceted approach that includes community-driven data collection, advanced annotation tools, and multi-modal integration. Beyond computational methods, interdisciplinary collaboration remains essential to develop robust and inclusive dialect recognition systems.


\subsection{Modeling Language Change and Evolution}

The evolution of low-resource languages is a dynamic and complex process. It is shaped by factors such as the shrinking or migration of communities that use the language over time, which leads to the simplification of grammatical rules or the integration of vocabulary from other languages. As a result, low-resource languages often develop unique dialects or variants. These languages typically face challenges in language technology applications due to limited corpora and insufficient technological support. However, with the rapid advancements in NLP, particularly with the advent of LLMs, new possibilities have emerged for tracking and analyzing the evolution of low-resource languages. These models not only facilitate the digital construction of language resources but also offer unprecedented tools to understand the dynamic changes of languages across time and space.

\subsubsection{The potential and limitations of LLMs in tracking the evolution of low-resource languages}

As one of the most advanced tools in the field of Natural Language Processing, large language models have demonstrated immense potential and advantages in tracking low-resource languages. However, this tracking process also faces several limitations due to factors such as data scarcity and unique grammatical structures. 

\textbf{Potential}

First, LLMs possess remarkable generalization abilities; after extensive training on large-scale multilingual data, they can adapt to low-resource languages through few-shot learning or fine-tuning on limited language samples. This adaptability makes them capable of recognizing and responding to the gradual changes within these languages. Moreover, self-supervised training methods endow LLMs with the ability to predict and generate language contextually, allowing them to infer emerging vocabulary and new expressions from minimal data samples.Additionally, LLMs can stay aligned with language evolution through regular parameter updates or periodic fine-tuning, ensuring their continued relevance to the latest linguistic trends in low-resource languages. 

 \textbf{Limitations}
 
The primary issue stems from the scarcity of available data, which limits the model’s exposure to comprehensive language patterns such as spoken versus written distinctions or context-specific expressions. Small sample sizes can lead to sampling bias, where the model may capture expressions specific to certain regions or communities, while overlooking broader linguistic diversity. Furthermore, the slow accumulation of low-resource language data delays the update process, hindering the model’s ability to reflect recent linguistic shifts accurately.

\subsubsection{Spatiotemporal variant processing in low-resource languages}

Low-resource languages often exhibit strong spatiotemporal variability, displaying significant linguistic differences across time periods, regions, and social groups. When applying LLMs to low-resource languages, it is crucial to consider this variability in model design and optimization. To address the spatiotemporal variations inherent in low-resource languages, several methods are commonly employed in the optimization and improvement of LLMs, including but not limited to the following approaches:

\textbf{Cross-lingual Transfer Learning}
Cross-lingual transfer learning, as referenced in \cite{cekinel2024cross}, is a widely used approach that leverages shared features from high-resource languages to aid in the processing of low-resource languages. This method helps capture fundamental linguistic features and can be generalized to accommodate different spatiotemporal variations.

\textbf{Removal of Explicit Language Tag Embeddings}
By removing explicit language classification encodings for words, this approach allows different languages to share feature representations\cite{conneau2019unsupervised}. It enables the model to benefit from data across multiple languages, thereby enhancing its broader linguistic understanding.

\textbf{Multimodal Architectures and Representation Learning}
Enhancing LLMs' ability to process low-resource languages through multimodal learning is an effective strategy\cite{mani2023large}. For example, combining visual and textual features can improve the model’s understanding ability of spatiotemporal variations and cultural context differences.

\textbf{Sentence Augmentation Techniques}
Some studies focus on sentence augmentation techniques to handle low-resource languages, such as Kazakh\cite{bimagambetova2023evaluating}. By generating more diverse sentences and expressions using LLMs, these methods address regional language variations. They enable the generation of more natural low-resource language sentences, expanding the data coverage and improving the model's generalization capabilities for new language inputs.

\subsection{Opportunities and Challenges}

Automated modeling provide linguistic scholar transformational opportunities with advancing research on low-resource languages like Quechua by enabling in-depth analysis of unique linguistic features and dialectal variability. For instance, it provides tools for interpreting sociolinguistic contexts, such as the placeholder "na," which functions both as a hesitation marker and as a substitute for omitted words, offering insights into bilingual code-switching and linguistic structure. Automated approaches also support the exploration of dialectal variability and linguistic borrowing\cite{languages9030082}. These contributions not only facilitate the preservation of linguistic diversity, but also enhance understanding of bilingual interactions and address the complexities inherent in dialectal variation.

However, significant challenges persist in achieving reliable and accurate modeling of dialects, with model errors often stemming from data shortages and standardization issues. The lack of high-quality annotated datasets remains a critical limitation, particularly for low-resource languages and under-represented dialects. This scarcity is exacerbated by inconsistencies between spoken and written forms, where variations in orthography, grammar, and writing style introduce additional complexities for model training and evaluation.

In conclusion, while LLMs provide promising solutions for handling diverse language variants, continued innovation in integrating dialect resources and addressing structural challenges is essential to achieve reliable and equitable linguistic modeling in low-resource settings.

\section{Applications of Low-Resource Languages in Historical Research}
Low-resource languages often carry rich historical and cultural information but are now at risk of gradual extinction. Protecting low-resource languages is not only a critical topic in linguistic research but also an essential aspect of historical studies. These languages have become “low-resource” due to a combination of factors. Over thousands of years, wars, colonialism, and political upheaval have forced certain language communities into displacement, leading to interruptions in language transmission \cite{saunt1999new,saunt2017our,saunt2020unworthy,saunt2005black,saunt2016age}. Industrialization and modernization have transformed traditional agricultural societies, sparking intense social change \cite{yang2021medicated}. Many people have migrated from rural areas to cities, abandoning their original languages in favor of more widely spoken ones. Under the influence of globalization, dominant cultures and languages often overshadow and displace weaker ones.

The importance of protecting low-resource languages within historical research cannot be underestimated. By preserving and studying these languages, we gain a deeper understanding of humanity’s past, enrich historical scholarship, and promote cultural diversity and social inclusivity. With the advancement of artificial intelligence and natural language processing technologies, efforts to protect low-resource languages have gained new momentum \cite{liu2023,liu2023summary,liu2024understanding,wang2024comprehensive}. Large language models and computer vision technologies can be used to identify, translate, and interpret low-resource language literature, providing powerful tools for historical research \cite{sommerschield2023machine}. The application of these technologies not only improves research efficiency but also expands the depth and scope of studies, allowing a broader range of low-resource language literature to be fully utilized. This section will delve into the application of LLMs in the interpretation of historical literature in low-resource languages.

\subsection{Interpretation of Historical Literature}
Literature in low-resource languages often carries rich historical and cultural information. Ancient texts, especially those written in low-resource languages, are important resources for studying specific cultures and historical periods. However, translating and interpreting these ancient texts currently face numerous challenges.

Specifically, these challenges include:

\textbf{Scarcity of Literature}: The quantity of ancient documents in low-resource languages is often limited and scattered across different locations such as archaeological sites, libraries, and private collections. Many of these texts are severely damaged due to their age, making portions of the content difficult to recognize.

\textbf{Linguistic Complexity}: The grammatical structures and vocabulary of low-resource languages differ significantly from modern languages, increasing the difficulty of translation. Even within the same language, there can be substantial differences in writing styles across different regions and historical periods.

\textbf{Scarcity of Language Experts}: The number of experts proficient in low-resource languages is limited. Many minority groups have integrated into modern civilized life and do not have mastery over their own ethnic languages.

There has already been some work using AI to translate and decode low-resource languages, such as the deciphering and restoration of the Dead Sea Scrolls \cite{popovic2021artificial,dhali2017digital,dhali2019binet,koopmans2024performance}, the recognition of ancient Egyptian hieroglyphs \cite{barucci2022ancient}, text restoration of ancient Greek inscriptions \cite{assael2022restoring}, the decoding of oracle bone script \cite{guan2024deciphering}, and the unified Visual-Linguistic understanding of oracle bone scripts\cite{jiang2024oraclesageunifiedvisuallinguisticunderstanding}. These early efforts, however, did not yet make use of LLMs. LLMs can play a pivotal role in addressing these challenges. Firstly, to combat the scarcity of literature, LLMs coupled with advanced computer vision techniques can assist in digitizing and reconstructing fragmented or damaged ancient texts. This not only preserves existing documents but also consolidates them into accessible digital archives for researchers worldwide. Secondly, regarding linguistic complexity, LLMs can be fine-tuned on available data, even if limited, to learn the unique grammatical structures and vocabulary of low-resource languages. Lastly, to mitigate the scarcity of language experts, LLMs can serve as virtual assistants, providing preliminary translations and analyses. This empowers a broader range of scholars and enthusiasts to engage with these languages, thereby expanding the pool of individuals who can contribute to their study and preservation.


\subsection{Data Mining and Knowledge Reconstruction in Historical Contexts}
LLMs have demonstrated remarkable potential in the mining of historical data and the restoration of knowledge, particularly concerning low-resource languages \cite{sommerschield2023machine}. These advanced models can learn the unique grammatical structures and vocabularies of these languages from limited datasets. By doing so, they can fill information gaps in historical documents and unearth unsystematized historical knowledge and informal records that have long been inaccessible to researchers.

Historical documents written in low-resource languages often suffer from significant information loss due to their age, poor preservation conditions, and the fragility of the materials used. These issues result in fragmented texts with substantial gaps, making it challenging to comprehend the complete historical context. LLMs can analyze the known portions of these texts, learning from their syntax and semantics to generate plausible reconstructions of the missing content. This process helps restore the integrity of historical documents, providing a more holistic view of the past.

Historical research has traditionally focused on formal documents like official records, treaties, and academic works. However, a wealth of historical information resides in informal records such as personal diaries, letters, folklore, and oral traditions. These sources offer invaluable insights into the daily lives, cultures, and social dynamics of historical communities but are often scattered across various archives and personal collections without systematic organization.

LLMs can process and analyze these informal records, extracting historical information that might otherwise remain hidden. By identifying patterns and drawing connections across disparate documents, LLMs provide new research perspectives and enable historians to construct more nuanced narratives \cite{liu2023summary}. This capability is particularly crucial for low-resource languages, where such informal records may be the primary sources of historical data.

In addition, LLMs can help to unify fragmented data from diverse sources, recognizing patterns that might otherwise be missed and aiding historians in constructing a more holistic view of historical events and cultural narratives. With advancements in technology, the application of LLMs in this field is expected to become more widespread and profound, enhancing historical analysis by making underutilized sources and scattered information accessible and meaningful. This trajectory points toward a future where LLMs enable deeper exploration of lost languages, unearth hidden knowledge, and enrich the field of historical studies in unprecedented ways.


\subsection{Challenges and Opportunities}
The application of LLMs to the historical study of low-resource languages has shown great potential, but still faces many technical difficulties in practice. Among them, the limitations of optical character recognition (OCR) technology, a key technology for digitizing ancient documents, are particularly prominent. The standard processing flow of OCR includes image preprocessing, text region localization, feature extraction, character recognition, and post-processing \cite{avyodri2022optical}. In these links, the accurate positioning of text region is crucial, which mainly involves two tasks of text detection and character segmentation.

The special characteristics of ancient documents make these two tasks more difficult to accomplish. Many documents are presented in handwritten or non-standardized fonts, and are often difficult to recognize due to blurring, fading, or physical damage \cite{memon2020handwritten}. Such conditions place high demands on OCR systems, making it difficult to effectively recognize areas containing text during text detection \cite{nguyen2021survey}. Similarly, some ancient documents (e.g., cursive scripts) have continuous writing and characters stacked on top of each other, which poses a challenge for character segmentation. These scenarios require the system to have higher resolution capability and accurate segmentation algorithms to avoid character misjudgment or character concatenation phenomenon.

Last but not least, there are many limitations in automatically converting ancient characters to modern languages. Low-resource languages often lack sufficient text corpora to provide high-quality training data for LLMs \cite{magueresse2020lowresourcelanguagesreviewpast}, which results in lower accuracy rates for translation or transcription of ancient characters. To further complicate matters, these languages tend to be morphologically variable and complex, and lack standardized writing rules. The morphology of a word may be varied and diverse depending on the context, tense, or syntactic roles, which poses a great problem for models to recognize and translate ancient texts. Finally, most of the current LLMs are primarily trained on large-scale datasets of modern languages, so researchers need to develop new approaches or adapt existing technical frameworks to address the particular challenges of low-resource scripts.

\section{Applications of Low-Resource Languages in Cultural Studies}

\subsection{Diversity of Cultural Corpora and Model Adaptation}

\subsubsection{Modeling of Low-Resource Languages in Cross-Cultural Contexts} 
Low-resource languages in cross-cultural contexts carry unique cultural values, often transmitting rich historical and social meanings through oral traditions, religious rituals, local ecologies, and artistic forms. Some oral traditions not only preserve the cultural semantics of ancestor worship but also shape intricate narrative rhythms. At the same time, the fusion of festive language with religious ritual language imbues regional languages with sacred elements in everyday usage. The integration of performing arts and architecture showcases the complex forms of language expression within cross-cultural settings\cite{erasmo2020theatre}. Low-resource languages are often deeply influenced by these cultural factors, leading to complex and diverse modes of expression. Therefore,  cross-cultural language modeling demands more than technical precision; it also requires integrating cultural understanding to develop models that reflect the cultural depth.

The corpora of low-resource languages, especially those carrying cultural histories such as oral traditions and folk narratives\cite{tehrani2023cultural}, are often insufficiently digitized or standardized, posing significant challenges for model adaptation. Some researchers have successfully incorporated these cultural differences into large models, yielding promising results\cite{li2024culturellm}. In the processing of low-resource languages, studies have shown that domain-adaptive pretraining (DAPT) can significantly enhance the performance of language models on culturally specific corpora\cite{meaney2024evaluating}.

\subsubsection{Challenges in Modeling Cultural Metaphors and Unique Usages}
Cultural metaphors and special usages pose unique challenges for LLMs. These linguistic phenomena are deeply embedded in specific cultural contexts, carrying rich historical, social, and emotional connotations. For instance, cultural metaphors often convey profound meanings through symbolic expressions, while special usages include slang, regional dialects, and cross-disciplinary terminology. Such linguistic features not only require models to understand context and surface meanings but also demand the ability to capture the complex cultural contexts and knowledge underlying the language.

To address this challenge, LLMs need more robust contextual modeling and knowledge integration capabilities. On the one hand, building a more diverse and culturally representative training corpus helps models encounter more instances of cultural metaphors and special usages. On the other hand, incorporating external knowledge bases or knowledge graphs can provide the model with richer background information and reasoning abilities. Additionally, techniques such as contrastive learning and few-shot learning can enhance the model’s generalization capabilities in low-resource domains, enabling it to understand and generate these linguistic features even with limited examples. Ultimately, developing culturally sensitive language modeling may improve cross-cultural communication and intelligent language interactions.

\subsection{Interaction Between Cultural Customs and Language Expression}

\subsubsection{How Language Models Understand Polysemy Across Cultural Contexts}

LLMs demonstrate a remarkable sensitivity and adaptability to cultural contexts when handling linguistic ambiguity. This is because the meaning of language is not only constrained by literal interpretation but is also deeply influenced by the cultural background and social norms in which it is used. Within the same language, specific words or expressions may exhibit different semantic layers due to cultural differences. By learning from vast, multilingual, and multicultural corpora, LLMs are capable of dynamically capturing contextual cues when analyzing text, enabling them to appropriately adapt to ambiguous words based on cultural context. This ability is particularly crucial in tasks such as language translation and cross-cultural dialogue generation, as it directly impacts the accuracy and cultural relevance of the generated language\cite{kibria-etal-2024-functional}.

However, Polysemy - the existence of multiple but related meanings for a single form - has always been problematic for purely structural accounts of meaning\cite{DEANE1988325}. LLMs face significant challenges when processing specific linguistic ambiguity - polysemy across different cultural contexts. The model's performance largely depends on the breadth and diversity of its training data. If certain cultural corpora are underrepresented, or if the semantic characteristics of a particular culture are overrepresented, the model may provide overly simplified interpretations of ambiguous words, overlooking the subtle cultural differences underlying the language. The proposed computational framework for quantifying polysemy, as outlined in recent research, introduces a novel approach to tackling this challenge. By combining graph-based methods with syntactic structures, such as dependency parsing and Ollivier Ricci curvature, this framework enhances the model’s capacity to estimate polysemy scores and map linguistic ambiguity with precision\cite{Goel2023BeyondTS}. Furthermore, linguistic ambiguity is often intertwined with metaphors, idioms, and other deep forms of expression, which poses even higher demands on the model. To enhance LLMs' generalization ability in multicultural contexts, it is essential to incorporate balanced and multi-layered corpus resources, along with cultural context annotations and domain-specific knowledge to guide the model's learning. These improvements can significantly enhance the model's adaptability in multicultural environments, making it more accurate and comprehensive in language generation and understanding tasks.

\subsubsection{Research on the Applicability of Language Models in Specific Cultural Scenarios}

LLMs have been applied globally, but their effectiveness in specific cultural contexts faces several challenges. These models are typically trained on vast datasets that encompass multiple languages and cultures. However, data biases and cultural specificities can lead to misunderstandings or misinterpretations in certain cultural settings. For instance, slang, proverbs, or traditional expressions that carry deep cultural significance in some languages may lose their original meaning or even be misunderstood in the content generated by the model. Additionally, metaphors, humor, and polite expressions unique to certain cultures might not align well with the model's general algorithms, limiting its interaction capabilities in complex cultural contexts.

To enhance the applicability of LLMs in specific cultural environments, researchers can approach the issue from three key areas: data, architecture, and evaluation. First, integrating more localized and high-quality datasets into model training can help cover a wider range of cultural backgrounds and expressions\cite{rao2024normad}. Second, model architectures can include targeted modules, such as cultural context embeddings or bias correction mechanisms, to better accommodate diverse cultural needs. Lastly, by collaborating with local users and cultural experts, a cultural sensitivity evaluation system can be established to thoroughly test the model’s performance in specific cultural contexts. These efforts would contribute to promoting the fair and efficient use of LLMs in cross-cultural communication, education, content generation, and other fields.

\subsection{Opportunities and Challenges}

\subsubsection{Opportunities: Role of Low-Resource Language Models}
Low-resource language models may have good potential in safeguarding endangered languages, traditions, and knowledge systems. These models empower the documentation and dissemination of oral traditions, folklore, and cultural narratives that might otherwise fade into obscurity due to a lack of balanced corpus resources. For example, the rich oral histories of indigenous communities, often passed down through generations, can be systematically transcribed and analyzed, ensuring their long-term preservation while simultaneously broadening their accessibility to a global audience. By bridging linguistic divides, language models\cite{Hutson2024} serve as conduits for cultural exchange, deepening cross-cultural intersectionality and fostering a sense of connection across diverse societies.

Beyond cultural preservation, low-resource language models offer a sophisticated method to unravel the complexities inherent in cross-cultural communication\cite{li2024cultureparkboostingcrossculturalunderstanding}. These models are particularly adept at addressing linguistic differences, such as dialectal variations, where they can act as multilingual experts, bridging and connecting diverse linguistic systems. With the potential to facilitate seamless translation and communication across languages and dialects without delays or loss of authenticity, language models are redefining the scope of cross-cultural interactions. By fine-tuning these models on culturally specific corpora, they can capture nuanced meanings with remarkable accuracy within the specific area, fostering a deeper understanding of the source culture while ensuring that translations and interpretations faithfully reflect original intention and the cultural depth. 

Furthermore, low-resource language models support in promoting cultural inclusivity by enabling underrepresented languages to present and participate in global discourse.  The preservation and update of low-resource languages in digital and educational spaces by language model, ensuring that these languages are not overshadowed by more dominant language. By integrating low-resource language models into cross-cultural research, scholars can uncover connections between linguistic structures and cultural customs, enriching the study of human diversity, and resolving linguistic inequality.

\subsubsection{Challenges: Misunderstandings and Bias within Low-Resource Language Models }
While the potential of low-resource language models in advancing cross-cultural communication and preserving endangered traditions is undeniably promising, several significant obstacles stand in the way.  A major challenge is the risk of misunderstandings and biases arising from insufficient cultural and contextual understanding. Currently, many language models are developed using training datasets that fail to fully encompass the linguistic diversity and cultural specificity of the communities they represents. As a result, the nuanced meanings of idiomatic expressions, metaphors, and culturally embedded terms are often misunderstood within cross-cultural context, leading to oversimplified translations or inaccuracies that fail to honor the richness of the original text\cite{liu2023wereafraidlanguagemodels}. 

Another pressing challenge lies in the inadvertent projection of dominant cultural perspectives onto minority languages or traditions. Many LLMs are predominantly trained on datasets drawn from high-resource languages and cultures, leading to a biased interpretative framework. This bias can dilute or distort the unique cultural and linguistic attributes of minority traditions, reducing their authenticity and misrepresenting their core meanings. Moreover, the scarcity of comprehensive resources for many low-resource languages exacerbates this issue. Without access to curated datasets and the necessary external contextual resources, LLMs struggle to accurately interpret and convey these elements. This challenge is particularly acute for oral traditions and languages with limited documentation, where much of the cultural essence remains undocumented or inaccessible\cite{kirk2023personalisationboundsrisktaxonomy}. Taking these and other unknown limitations into consideration, there is a long way to achieve the effective application for low-resource language models in cross-cultural context.


\section{Applications in Literary and Religious Studies}
\subsection{Automated Translation and Analysis of Low-Resource Literary Works}
Literary works play a crucial role in the language system. Literary works are not only the simple transmission and preservation of information, but they also carry the author's emotions, thoughts, and even contain the cultural background of a region or even a country\cite{olsen1982meaning}. Therefore, automated translation and analysis of literary works is an important way to understand and inherit a language\cite{karabayeva2024evaluating}. The translation of literary works is not just about simple word conversion, but about conveying the same emotions and meanings as the original work in a new language environment. Similarly, the analysis of literary works also needs to uncover the deep hidden meanings of the works, rather than simply retelling and summarizing them. For low-resource languages such as Croatian and most other less used languages\cite{dundjer2020automatic}, it is crucial to use LLMs to translate and analyze literary works for understanding, protecting, and even cultural heritage.

However, using LLMs to translate and analyze literary works in low-resource languages is a relatively difficult process. This is because for low-resource languages, there is usually a lack of large-scale parallel corpora or high-quality datasets\cite{lin2020towards}, which are the basis for LLMs to understand low-resource languages, making it difficult to process literary works. Secondly, literary works in low-resource languages often contain unique cultural backgrounds and local characteristics. Without sufficient cultural background knowledge, it is difficult for LLMs to accurately translate these unique cultural elements. To address these challenges, researchers have proposed various methods, including building corpora of low resource languages, cross language transfer learning\cite{huang2020cross}, and using professionals to proofread and revise the results provided by large language models.

Another key issue in the automated translation and analysis of low-resource literary works using LLMs is whether they can accurately identify the styles of different literary works. Different writers have different writing styles, which often reflect their unique perspectives and artistic pursuits. In order to better capture and preserve these styles, LLMs must have higher adaptability. Specifically, fine-tuning can be made to the LLMs to capture the stylistic characteristics of specific writers, such as romanticism, realism, etc., achieving more natural and fluent translation and analysis.

\subsection{Identification of Specific Vocabulary and Concepts in Religious Texts}
Religious texts are usually written in low-resource languages and are important repository of theological, philosophical, and cultural knowledge. They are renowned for their unique terminology, symbolic structures, and contextual meanings\cite{agliz2015translation}, and can be better studied and analyzed using LLMs.

Lai's research\cite{lai-etal-2023-turn} offers an interdisciplinary approach to linguistics, combining qualitative theory (Dialogic Syntax, Construction Grammar) with quantitative computational analysis focused on corpus linguistics, statistical modeling, and information theory. This "usage-based" perspective views language as a dynamic system emerging from human interaction, exemplified by the analysis of texts like the Diamond Sutra. He argues that highly structured, non-spontaneous texts, like orally transmitted religious texts, provide a valuable, high-signal source for understanding linguistic patterns, suggesting a novel approach to LLM training by curating stylized, ritualized texts for enhanced structural and relational richness.

Religious texts generally have specific language and structural attributes. Many low-resource religious texts use vocabulary with profound theological significance, often lacking direct equivalents in modern language, which are referred to as sacred terminology. For example, the Sanskrit term "Nirvana" \cite{pasadika2007nirvana}represents a complex concept that differs greatly between Hinduism and Buddhism, including liberation, cessation, and transcendence; In Christian texts, Greek words like "agape" and "logos" are filled with theological and philosophical meanings. In addition, some religious texts may use repetitive phrases for ritual, memory, or meditation purposes, and these repetitions need to be explained and preserved during the analysis and translation process\cite{elewa2014features}. Secondly, many religious texts heavily rely on symbolism and fables\cite{jakel2002hypotheses}. For example, the recurring theme of light in sacred texts such as Vedic hymns or the Bible conveys a range of meanings from divine existence to enlightenment. Finally, religious texts are deeply rooted in their social and historical context. Words, phrases, and even entire paragraphs can reflect specific historical events, cultural norms, or local belief systems, and require contextual understanding to be accurately interpreted.

The LLMs provide revolutionary possibilities\cite{Sacred_Texts_Digital_Age} for the study of religious texts. In cross religious contexts, LLMs can compare and study traditional texts of different religions, identify terminology and doctrines. For example, LLMs can compare the concept of "sympathy" expressed in Buddhist Pali scriptures, Christian New Testament texts, and Islamic hadith, providing insights into both general and specific traditional interpretations. LLMs can help explain the subtle differences of the same term in different religious backgrounds. For example, the Sanskrit term "dharma" \cite{brockington2004concept} in Hinduism may refer to justice or responsibility, while in Buddhism, it means the teachings of the Buddha or universal truth. LLMs trained extensively in context can help eliminate ambiguity in these terms during analysis or translation processes. In the cross-religious context, the same symbol may have different meanings in different traditions. Religious and philosophical texts are not simply repositories of information; they are highly structured performances of knowledge and authority, featuring unique rhetorical and structural patterns that are not present in everyday communication. A simple factual summary of such texts fails to capture their underlying rhetorical and dialogic function. For example, the lion is a symbol of strength and divinity in the Judeo Christian tradition, while in the Buddhist tradition it represents wisdom and courage. LLMs can compare these symbols in the corpus and enrich the interpretation by placing them in their respective traditions.

\subsection{Opportunities and Challenges}

\subsubsection{Opportunities: Innovative Applications of Language Models}
The application of language models to low-resource literary and religious texts presents remarkable opportunities for preserving and revitalizing religious heritage. These models serve as vital bridges, reducing barriers to literary appreciation while making classic literary works more accessible to contemporary audiences. By delving into texts from low-resource languages such as Croatian and Quechua, language models can help identify cultural nuances, historical context, and emotional aspects embedded within these works, supporting broader engagement and deeper appreciation.

To revive low-resource literary works and ensure their resonance with modern audiences, particularly younger generations, the presentation and dissemination of these literary works must align with contemporary tastes and cultural dynamics. Language models can facilitate this by economically producing accessible, high-quality, and readable content tailored to the preferences of target audiences. Leveraging popular platforms and modern formats, such as interactive media or engaging digital narratives, can reintroduce overlooked literary treasures in a way that is both appealing and relevant. This approach not only preserves the vibrancy and influence of these cultural artifacts but also fosters a deeper understanding of global heritage while enabling meaningful dialogue on religious and cultural complexities. Specifically, it helps younger readers engage more effectively with religious texts, improving theological comprehension\cite{leblanc2012ancient} and contributing to the movement of intercultural peace. 

Beyond accessibility and revitalization, LLMs can also be a powerful tool for academic and theological inquiry. In the realm of religious studies, language models offer a unique opportunity to address doctrinal conflicts and enrich theological discourse in several areas. By analyzing religious texts, these models can uncover common principles and provide clarity in areas where doctrinal interpretations have historically diverged. From a technological perspective, the systematically machine-learnable structure inherent in the Buddha-Subhti dialogue, as developed by Ryan Lai, presents several contributions to the fields of AI\cite{lai-2024-tupleised} and low-resource language processing. These contributions include the comprehensive modeling of epistemic status, the operationalization of rhetorical routines, and the recognition of the inherent value of stylized corpora for religious text, which furnish high-quality training data due to their structured textual nature. Through a profound understanding of linguistic routines and their associated social functions, this research endeavors to facilitate the development of more nuanced and communicatively proficient LLMs, particularly within intricate and specialized domains.

This article, alongside that of scholars like Carolyn Medine, underscores a crucial point: the successful application of LLMs in religious studies requires a comprehensive understanding of both the linguistic structure of sacred texts and their broader social and cultural contexts. For example, a purely textual analysis of Buddhism is insufficient without considering how entrenched racial and cultural dynamics shape its interpretations, as examined in \cite{yancy2019buddhism}. Likewise, Carolyn Medine investigates the interdisciplinary culture of race, discrimination, and religion\cite{Medine_2022} in her tribute article to bell hooks, a black feminist and black buddhist, it highlights the need for LLMs to incorporate the nuances of race, gender, and social context in their analysis beyond a purely textual approach. Therefore, these models are capable of adapting and refining doctrinal interpretations to align with evolving perspectives, while remaining anchored in traditional frameworks. A notable application is the preservation and dissemination of religious traditions tied to specific cultural contexts, such as African-American theology\cite{jones_medine_2023}. With thoughtful development and ethical oversight, these models have the capacity to establish balanced frameworks that address doctrinal ambiguities while expanding the accessibility of religious knowledge across diverse cultural boundaries.

\subsubsection{Challenges: Limitations in Models’ Understanding of Implicit Meaning}
While there are multiple innovative applications of language models in low-resource literary and religious texts, a critical challenge in their application to ancient literature and religious texts lies in their limited capacity to interpret the implicit and multilayered meanings. Such texts often draw on intricate metaphors, allegories, and symbolic references that demand not just linguistic interpretability but also cultural, historical, and philosophical insight. Existing language models, grounded in patterns derived from existing datasets, inevitably lack the depth and nuance required to faithfully interpret these complexities, and professional human translations\cite{zhang2024goodllmsliterarytranslation} consistently outperform LLMs translations in certain contexts.

The limitation\cite{Mahowald_2024} is compounded by the scarcity of preserved external resources and relevant data for many low-resource languages and traditions. While some texts have survived in written form, much of the oral tradition, cultural knowledge, and historical context that breathe life into these works have been lost to time. In particular, religious and philosophical texts are highly structured performances of knowledge and authority, featuring unique rhetorical and structural patterns that are not present in everyday communication. A shallow factual summary of such texts fails to capture their underlying rhetorical and dialogic function. Without these necessary components, language models struggle to discern the deeper connections within the texts, frequently producing interpretations that are either reductive or outright erroneous. Meanwhile, LLMs often exhibit biases and cultural misrepresentations, as revealed by debate-induced evaluations, which highlight their tendency to align with dominant linguistic or cultural perspectives in training data\cite{demidova-etal-2024-john}, underscoring the need for fairness and contextual sensitivity in multilingual training. 

To overcome these limitations, there is an urgent need for curated datasets enriched by expertise within fields such as history, philosophy, and cultural studies. These datasets must not only provide lexical and grammatical inputs but also encapsulate the symbolic, contextual, and philosophical depth of the source material. Furthermore, fostering collaboration between scholars from diverse disciplines can help bridge the gap between raw textual data and the broader interpretative frameworks required to fully understand these works. 


\section{Technical and Ethical Considerations for Large Language Models in Low-Resource Languages}
\subsection{Technical Challenges}
LLMs face significant challenges when applied to low-resource languages due to a variety of factors, primarily stemming from the scarcity and quality of available data~\cite{kholodna2024llms,wang2024improving,ogueji2021small,hedderich2020survey}. Unlike high-resource languages such as English or Chinese, low-resource languages often lack the extensive and standardized datasets required for effective training. Available data is typically fragmented, non-standardized, or entirely absent, making it difficult for LLMs to learn nuanced linguistic patterns. Furthermore, critical NLP tools, such as part-of-speech taggers and annotated datasets, are frequently unavailable for these languages, further impeding the development of robust models.

Another pressing issue is the lack of computational resources in regions where low-resource languages are predominantly spoken~\cite{hedderich2020survey,hasan2024large}. Training large-scale models demands significant computational power and infrastructure, which are often inaccessible in resource-constrained environments. The high costs associated with training and deploying these models exacerbate the problem, limiting the ability of local researchers and developers to create and utilize advanced NLP tools tailored to their linguistic and cultural needs.

The linguistic and cultural diversity inherent in low-resource languages presents additional hurdles. For example, Africa alone is home to over 2,000 languages, each with unique grammatical rules, vocabulary, and cultural contexts~\cite{tonja2024inkubalm}. This diversity complicates the development of multilingual models capable of performing consistently across languages. Moreover, many low-resource languages lack sufficient digital representation, resulting in inadequate data for pretraining and instruction tuning. Consequently, LLMs trained on such data often underperform, reflecting the foundational gap in resources necessary for their development.

Low-resource languages also face specific challenges during training due to inefficiencies in tokenization processes. For instance, non-Latin scripts, such as Bengali, are often over-tokenized by standard methods like Byte Pair Encoding (BPE)~\cite{mahfuz2024too}. Over-tokenization leads to higher computational costs and lower information density, negatively impacting model efficiency and performance. These inefficiencies highlight the need for more refined tokenization methods tailored to the characteristics of low-resource languages.

To further complicate matters, the data that is available for these languages is often derived from machine translation, introducing biases and inconsistencies into training and evaluation. This reliance on imperfect data sources undermines the reliability of models and hampers their ability to generalize effectively. In cases where pretraining, instruction-tuning, or reinforcement learning with human feedback (RLHF) datasets are available, their quality and coverage are often insufficient to meet the needs of robust language modeling.

In summary, the challenges of applying LLMs to low-resource languages are multifaceted, encompassing issues of data scarcity, computational resource limitations, linguistic diversity, tokenization inefficiencies, and reliance on imperfect data sources. Addressing these challenges requires concerted efforts to improve dataset availability and quality, develop computationally efficient methods, and design culturally and linguistically sensitive models.

\subsection{Ethical Challenges}
Ethical and fairness concerns also arise when applying LLMs to low-resource languages. Models trained on limited and biased datasets may perpetuate existing linguistic or cultural biases, leading to inaccurate or inappropriate outputs. These biases can negatively impact the fair application of NLP tools in diverse cultural contexts. For instance, Low-resource Indigenous languages often encode culturally sensitive knowledge integral to their communities' identity and heritage. Indiscriminate harvesting of Indigenous language data from online sources poses significant risks to privacy and cultural sovereignty. For example, much of the digital text available for Indigenous languages, such as South Sámi, may come from private or semi-private community
contexts~\cite{wiechetek2024ethical}. This makes uncritical use of such corpora problematic, as it could expose knowledge the community considers sacred or confidential. Ethical language technology development requires building trust with language communities, ensuring data use respects cultural boundaries and adheres to explicit agreements on data utilization.

LLMs, while promising, can inadvertently worsen the marginalization of low-resource languages. Low-resource Indigenous languages often have limited digitized resources, leading to minimal and potentially flawed representation in language models. Uncorrected errors in machine-generated outputs risk amplification in subsequent models, creating a feedback loop that distorts authentic language usage ~\cite{wiechetek2024ethical}. Furthermore, the dominance of high-resource languages in training corpora can lead to linguistic homogenization, further weakening the already vulnerable position of low-resource languages. This imbalance threatens the linguistic diversity and the preservation of cultural identities.

By prioritizing quantity over quality, LLMs risk imposing majority language norms on Indigenous languages, eroding their distinct grammatical structures, idiomatic expressions, and cultural relevance. Addressing this issue demands a deliberate effort to include native speakers and language experts in the development process, ensuring accurate and equitable representation of Indigenous languages.

These challenges highlight the importance of ethical practices in developing language technologies for Indigenous communities, focusing on protecting cultural knowledge and promoting linguistic diversity in a way that empowers rather than marginalizes these languages.

\subsection{Policy Recommendations}
Promoting ethical and impactful applications of NLP for low-resource Indigenous languages requires the adoption of thoughtful and inclusive policies. These policies must respect the linguistic and cultural integrity of these communities, prioritize equitable partnerships, and focus on fostering the sustainable development of linguistic resources. Encouraging open data sharing, advocating for equitable research practices, and safeguarding the linguistic ecosystem are essential measures for advancing NLP capabilities for Indigenous languages. Such protection policies not only address the practical challenges of dataset development but also ensure that the resulting tools align with the long-term interests and aspirations of the communities they aim to support. By prioritizing ethical collaboration and cultural preservation, stakeholders can empower Indigenous language communities to thrive and maintain their heritage in the digital age.

\textbf{Encouraging Open Dataset Sharing and International Collaboration}
Low-resource Indigenous language communities often face challenges due to the lack of extensive literary traditions, ongoing standardization processes, and inconsistencies in written texts. Many texts available digitally exhibit significant variations in spelling and grammar, often deviating from the norms desired by the language community. Additionally, a notable proportion of these texts are authored by non-native speakers or language learners, resulting in corpora that are not representative of the language's authentic usage. Governments and institutions should prioritize open dataset sharing initiatives and international collaboration. These collaborative efforts can ensure datasets are curated, verified, and annotated by native speakers and language experts. Such partnerships will facilitate the development of high-quality, representative corpora that align with the community's cultural and linguistic standards. By promoting international cooperation, organizations can pool resources, expertise, and perspectives to advance the preservation and revitalization of these languages.

\textbf{Advocating for Equitable Research Policies and Language Ecology Preservation}
The development of ethical and effective machine learning-based NLP tools requires policies that center the needs and voices of low-resource Indigenous language communities. Outputs from these tools should undergo rigorous evaluation by language experts from within these communities, ensuring cultural and linguistic accuracy. Clear authorship verification and respect for data ownership are critical to maintaining trust and preventing the exploitation of Indigenous knowledge. Policies must advocate for research practices that treat Indigenous communities as equal partners rather than subjects. This involves ensuring that tools developed using Indigenous language data are freely available to the community and modifiable to meet their unique needs. Furthermore, these policies should emphasize acknowledging and crediting the contributions of data providers and actively preserving the linguistic ecology. This includes respecting the cultural, economic, and ideological value of original language data and ensuring that it is used in ways that support community-driven goals.

\section{Foundational Framework for LLMs in Low-Resource Language Research}

\subsection{Definitions and Types of Low-Resource Languages}

A \textbf{low-resource language} refers to a language that suffers from a significant scarcity of linguistic resources, such as corpora, dictionaries, and annotated datasets. These languages often lack large-scale digital support, posing challenges for NLP and computational linguistic research~\cite{Hedderich2020ASO}. Despite their diversity and cultural significance, low-resource languages are underrepresented in modern technological and computational advancements. Below, we delve into the key classifications of low-resource languages, their unique challenges, and the role of LLMs in addressing these gaps.

\subsubsection{Dialects}

Dialects are regional or social varieties of languages that often lack standardized written forms. They are predominantly used in oral communication and are rich in linguistic diversity. For example, Chinese Cantonese, Hokkien, and Shanghainese are widely spoken but remain underrepresented in computational linguistics due to the absence of robust, large-scale corpora.

The challenges in processing dialects include:
\begin{itemize}
    \item \textbf{Diversity}: Dialects often exhibit significant phonological, lexical, and syntactic differences, even within the same language family.
    \item \textbf{Resource scarcity}: Dialects are rarely documented comprehensively in written form, and existing data are typically fragmented.
    \item \textbf{Corpus complexity}: Capturing the linguistic rules and structures of dialects in a corpus requires extensive resources, including phonetic annotations and regional usage patterns.
\end{itemize}

Large language models, when augmented with transfer learning techniques or pre-trained on related high-resource languages, have the potential to bridge these gaps. However, dialects' inherent variability still presents a major obstacle to achieving high performance in NLP applications~\cite{LSSZ202408005}.

\subsubsection{Ancient Languages}

Ancient languages, such as Latin, Sanskrit, and Ancient Greek, hold immense historical and cultural significance but are no longer in widespread use for daily communication. These languages pose unique computational challenges:
\begin{itemize}
    \item \textbf{Limited and fragmented corpora}: Ancient texts often exist in non-standardized formats or incomplete manuscripts.
    \item \textbf{Lack of alignment with modern linguistic paradigms}: Ancient languages frequently exhibit archaic grammatical structures and vocabulary, requiring specialized models for processing.
    \item \textbf{Preservation and digitization}: Many ancient scripts lack comprehensive digitization, hindering the availability of training data for machine learning models.
\end{itemize}

For instance, while Latin remains a cornerstone of religious and academic studies, its scarcity of annotated datasets and linguistic ambiguities complicate NLP tasks such as translation, semantic parsing, and automated analysis of ancient texts. Large language models designed with symbolic reasoning capabilities and fine-tuned on ancient text corpora may offer promising avenues for this domain.

\subsubsection{Endangered Languages}

Endangered languages are at risk of extinction, often spoken by small, isolated communities. Many indigenous languages in regions such as Australia, Africa, and the Americas face dwindling numbers of speakers, limited geographical reach, and minimal documentation. Cherokee language evaluation through the ChrEn dataset reveals best-case performance of 15.8/12.7 BLEU for in-domain translation, with only 2,000 fluent first-language speakers remaining. The Cherokee Nation's loss of 346 fluent speakers between 2020-2022 underscores the urgency of developing effective NLP tools~\cite{zhang2022cherokee}.

The challenges in endangered language processing include:
\begin{itemize}
    \item \textbf{Speaker scarcity}: Fewer native speakers mean fewer opportunities for collecting spoken and written data.
    \item \textbf{Minimal academic support}: Endangered languages often lack institutional resources for linguistic preservation.
    \item \textbf{Urgency}: The rapid decline of speakers demands accelerated efforts for documentation and computational support.
    \item \textbf{Performance limitations}: For instance, Quechua~\cite{chen2024queen}, even specialized models like QueEn (2024) achieve only 17.6 BLEU compared to 1.5 for standard GPT models.
\end{itemize}

To address these challenges, LLMs can leverage \textit{transfer learning}, \textit{zero-shot learning}, and \textit{unsupervised methods} to process limited data effectively. Furthermore, community-driven initiatives can play a pivotal role in collecting and annotating linguistic data.

\subsection{Current State of Foundation Models for Low-Resource Languages}

As of August 2025, the landscape of LLMs has evolved significantly with the confirmed releases of \textbf{GPT-5} (OpenAI, August 7, 2025), \textbf{Gemini 2.5 Pro} (Google, March-July 2025), and \textbf{Claude 4} (Anthropic, May 2025). GPT-5 shows performance improvements with 94.6\% accuracy on AIME 2025 and 74.9\% on SWE-bench Verified, featuring unified reasoning architecture and approximately 45\% fewer factual errors than GPT-4o~\cite{openai2025gpt5}. Gemini 2.5 Pro introduces adaptive reasoning budgets with ``thinking'' capabilities and native multimodality across text, images, audio, and video with 1M+ token context windows~\cite{google2025gemini}. Claude Sonnet 4 offers hybrid reasoning models with 200K context windows standard and 1M tokens available in beta, achieving 96\% accuracy across 12 major languages~\cite{anthropic2025claude4}.

Despite these advances, a fundamental disparity persists between high-resource and low-resource language performance. The Stanford HAI White Paper~\cite{stanford2025hai} identifies that "most major LLMs underperform for non-English—and especially low-resource—languages; are not attuned to relevant cultural contexts; and are not accessible in parts of the Global South". This gap becomes particularly pronounced when examining specific low-resource languages, where even state-of-the-art models frequently perform below random baselines.

\subsection{Scarcity and Complexity of Corpora in Low-Resource Languages}

Low-resource languages are characterized by the scarcity of corpora, which significantly hinders their integration into modern NLP systems~\cite{LSSZ202408005}. Unlike high-resource languages such as English and French, which have benefited from decades of extensive corpus development, these languages often lack sufficient datasets. The challenges in data collection are multifaceted, involving logistical difficulties, financial constraints, and cultural sensitivities. Assembling even basic corpora for such languages can be a significant undertaking, and the lack of resources limits the development of NLP models tailored to these linguistic contexts.

The available datasets for low-resource languages often suffer from poor quality and limited scope, making them suboptimal for robust model training~\cite{ranathunga2023neural}. Inconsistencies within the datasets and the absence of comprehensive linguistic coverage further restrict their utility. The MiLiC-Eval ~\cite{zhang2025milic} benchmark reveals that Tibetan requires 70-430 tokens per sentence compared to 27 for English, highlighting fundamental architectural limitations in handling morphologically complex scripts.

The problem is exacerbated by the underrepresentation of these languages in AI research, as commercial interests have historically focused on high-resource languages. This imbalance has resulted in technological advancements that predominantly benefit languages with larger speaker bases and more substantial economic influence.

The linguistic complexity of low-resource languages further amplifies these challenges. Many of these languages possess unique grammatical structures, such as the lack of clear distinctions between grammatical categories or the use of non-linear syntax~\cite{ranathunga2023neural}. These features make it difficult to adapt existing NLP techniques designed for high-resource languages. Moreover, orthographic diversity adds another layer of difficulty. Some languages lack standardized writing systems, while others use scripts that are challenging to digitize or process with current computational tools.

Additionally, regional variations within low-resource languages pose a significant obstacle to corpus development. Differences in pronunciation, vocabulary, and syntax across geographical areas require highly localized datasets to capture the full linguistic diversity of these languages~\cite{LSSZ202408005}. Such regional variations demand substantial effort to develop representative and contextually accurate corpora. These factors collectively underscore the pressing need for innovative strategies to overcome the intertwined issues of scarcity and complexity in low-resource language processing.

\subsection{Role of Large Language Models in Addressing Challenges}
Recent research has crystallized around three primary objectives for advancing low-resource language support in foundation models. First, bridging the performance gap between high-resource and low-resource languages through innovative architectural and training approaches. Second, preserving linguistic diversity and cultural heritage through community-centered development methodologies. Third, developing evaluation frameworks that accurately capture the unique challenges of morphologically complex and typologically diverse languages.

Zhang et al. established fundamental principles and proposing machine-in-the-loop processing approaches that prioritize community engagement ~\cite{zhang2022cherokee}. This work demonstrates that successful low-resource language NLP requires not merely technical innovation but collaborative frameworks that respect indigenous data sovereignty and cultural contexts. The TLUE benchmark for Tibetan ~\cite{gao2025tlue}, which reveals that most current LLMs including GPT-4 perform below the 25\% random baseline on Tibetan tasks, with Claude-3.5-Sonnet achieving the highest accuracy at only 35.6\%. Similarly, the AmericasNLI dataset ~\cite{ebrahimi-2022-americasnli} covering ten Indigenous languages of the Americas shows that XLM-R's zero-shot performance averages merely 38.48\% accuracy, only slightly above random chance.

While the road ahead is challenging, the integration of community efforts, advanced computational methods, and interdisciplinary research holds the potential to unlock the linguistic treasures embedded within low-resource languages. These advancements will not only benefit NLP but also contribute to the preservation of linguistic and cultural diversity.

\subsection{Techniques Supporting Low-Resource Languages}

To improve the performance of LLMs in processing low-resource languages, researchers have proposed a range of technical approaches. These methods aim to enhance the adaptability of LLMs by improving training data quality, leveraging cross-language transfer, and incorporating multi-modal information. Together, these techniques contribute to significant advancements in processing low-resource languages, despite their inherent challenges.

\subsubsection{Transfer Learning}

\textbf{Transfer learning} is a widely adopted technique to address the challenges posed by low-resource languages~\cite{zoph2016transfer}. By pre-training a model on high-resource languages and fine-tuning it on low-resource languages, the model can acquire foundational linguistic knowledge from the former and adapt to the unique characteristics of the latter. 

\textbf{Recent advances in chain-of-thought reasoning have revolutionized transfer learning for low-resource languages.} Research demonstrates that scaling inference-time compute through CoT reasoning enables models to leverage high-resource language knowledge more effectively~\cite{yeo2025longcot}. Fine-tuning on just 1,000 reasoning traces in languages like Swahili produces 33.8\% performance gains, as the model learns to use well-represented languages as internal pivot points during reasoning. This represents a paradigm shift from static knowledge transfer to dynamic, reasoning-based transfer where the model explicitly generates intermediate steps in high-resource languages before producing low-resource language outputs.

Key benefits of transfer learning include:
\begin{itemize}
    \item \textbf{Knowledge reuse}: High-resource languages serve as a knowledge base, enabling the model to understand basic linguistic structures, syntax, and semantics of low-resource languages.
    \item \textbf{Resource efficiency}: Reduces the amount of data required to train the models specifically for low-resource languages.
    \item \textbf{Improved generalization}: Enhances model robustness in handling sparse or noisy data.
\end{itemize}

For example, pre-trained language models such as BERT, GPT, and RoBERTa have been shown to adapt effectively to new languages and domains through fine-tuning, improving their performance in low-resource settings. However, the effectiveness of transfer learning often depends on the degree of linguistic similarity between high-resource languages and low-resource languages.

\subsubsection{Cross-Language Pretraining}

\textbf{Cross-language pretraining} has evolved from early multilingual models to sophisticated architectures that better handle linguistic diversity. Initial approaches, exemplified by models like mBERT, XLM, and XLM-R~\cite{gonen2020s}, train on multilingual corpora to learn shared representations across languages.

These multilingual models offer several advantages:
\begin{itemize}
    \item \textbf{Shared representations}: Encode universal linguistic features, facilitating the transfer of cross-linguistic knowledge.
    \item \textbf{Scalability}: Support multiple languages without extensive language-specific fine-tuning.
    \item \textbf{Zero-shot capabilities}: Enable performance on unseen languages by leveraging related language knowledge.
\end{itemize}

\textbf{However, a fundamental limitation of dense multilingual models is parameter interference}—all languages compete for the same parameter space, often resulting in high-resource languages dominating the representations. The emergence of sparse Mixture of Experts (MoE) architectures directly addresses this challenge~\cite{zhou2022moe}. Unlike dense models, MoE maintains dedicated expert networks for different language families, preventing parameter interference while preserving the benefits of shared learning.

In practice, MoE architectures have achieved 2.5 BLEU point improvements in zero-shot translation scenarios. The gating networks learn to route queries based on linguistic features rather than explicit language identification, enabling nuanced cross-linguistic knowledge sharing. For instance, mBERT's effectiveness in cross-language question answering is now enhanced by MoE's ability to maintain language-specific expertise while leveraging shared multilingual embeddings.




\subsubsection{Multi-Task Learning}

\textbf{Multi-task learning} involves training models on multiple related tasks simultaneously, allowing them to share knowledge across tasks~\cite{9392366}. This approach is particularly beneficial for low-resource languages as it enables the model to leverage auxiliary tasks to improve performance. 

Traditional multi-task learning benefits include shared knowledge across complementary tasks, regularization effects that prevent overfitting, and efficient resource usage through unified training frameworks. Tasks such as translation, sentiment analysis, and text classification provide complementary signals that enhance model generalizability for low-resource languages.

The introduction of \textbf{dynamic routing mechanisms}, particularly in GPT-5's unified architecture, has elevated these benefits to new levels~\cite{openai2025gpt5}. Rather than treating all tasks equally, the model now employs a real-time router that dynamically allocates computational resources based on task complexity and linguistic demands. For low-resource languages, this means:

\begin{itemize}
    \item \textbf{Shared knowledge}: Tasks such as translation, sentiment analysis, and text classification provide complementary information, enhancing model generalizability.
    \item \textbf{Regularization}: Multi-task learning acts as a form of regularization, preventing overfitting on small datasets.
    \item \textbf{Efficient resource usage}: Combines multiple objectives into a unified training framework, reducing the need for extensive single-task datasets.
    \item \textbf{Continuous learning}: The router learns from user interactions and measured correctness, progressively improving its handling of low-resource language patterns.
    \item \textbf{Task-aware optimization}: Different aspects of language processing (syntax, semantics, pragmatics) can be weighted differently based on the specific challenges of each low-resource language.
\end{itemize}

By exposing the model to diverse tasks, multi-task learning helps the model generalize better to low-resource language challenges, including those with limited linguistic features.





\subsubsection{Data Augmentation}

\textbf{Data augmentation} addresses the problem of insufficient training data by generating additional examples and leveraging available resources more effectively~\cite{ragni2014data}. This technique enriches datasets through both preprocessing and inference-time strategies, providing models with more diverse and representative samples.

Common data augmentation methods include:
\begin{itemize}
    \item \textbf{Synthetic data generation}: Machine translation creates parallel corpora by translating high-resource language datasets into low-resource languages. Back-translation has proven particularly effective in machine translation and text classification tasks, significantly boosting model performance.
    \item \textbf{Paraphrasing}: Generating alternative expressions for existing sentences increases linguistic diversity and helps models learn invariant representations across different phrasings.
    \item \textbf{Noise injection}: Adding controlled perturbations to data simulates real-world variability and improves model robustness to input variations.
    \item \textbf{Dynamic resource integration}: LLMs like GPT-5 and Claude 4 can alternate between reasoning and tool use during inference, incorporating external resources such as dictionaries and parallel corpora as needed~\cite{openai2025gpt5,anthropic2025claude4}. This enables real-time augmentation where the model searches for and integrates relevant linguistic resources based on specific query requirements.
\end{itemize}

Expanded context windows have fundamentally changed how augmentation can be applied. With GPT-5 supporting 400K tokens and Claude Sonnet 4 supporting 1M tokens~\cite{openai2025gpt5,anthropic2025claude4}, entire low-resource language corpora can be provided as context during inference. This capability enables several key advantages: First, for many endangered languages, the complete written corpus fits within a single context window, effectively providing the model with all available knowledge at inference time. Second, models can become temporarily specialized for specific languages by loading relevant resources into context, eliminating the need for separate fine-tuning. Third, historical texts and oral tradition transcriptions can be used directly without risk of corruption through synthetic generation processes.

The combination of synthetic augmentation and inference-time resource integration provides complementary benefits. While synthetic methods expand training data coverage before deployment, dynamic integration allows models to adapt to specific needs during use. For low-resource languages, this dual approach is particularly valuable—static augmentation provides baseline coverage while dynamic augmentation addresses specialized or unexpected queries that arise during deployment.

These augmentation strategies must be carefully calibrated for each language's specific characteristics. Languages with related high-resource neighbors benefit more from translation-based augmentation, while linguistically isolated languages may rely more heavily on context-based approaches and paraphrasing within the available data.

\subsubsection{Multi-Modal Integration}

Incorporating \textbf{multi-modal information}, such as audio, video, and images, has emerged as a promising approach to enhance low-resource language processing. By leveraging non-textual data, models can better understand and represent the contextual nuances of languages.

Advantages of multi-modal integration include:
\begin{itemize}
    \item \textbf{Complementary information}: Audio and visual data provide cues that are absent in textual data, such as pronunciation and gestures.
    \item \textbf{Improved accessibility}: Enables the processing of languages with oral traditions or limited written records.
    \item \textbf{Enhanced semantic understanding}: Multi-modal models can capture richer, context-aware representations.
\end{itemize}

For instance, speech-to-text and text-to-speech models trained on multi-modal data have shown significant improvements in processing spoken languages with limited textual resources.

\subsection{Adaptation and Gaps Between LLMs and Low-Resource Languages}

In recent years, LLMs have made significant progress and demonstrated immense potential in the field of humanities research. These models can process vast amounts of text, uncover hidden information, and provide new perspectives and methods for tasks such as ancient text interpretation and cultural analysis. This capability is particularly beneficial for humanities research, especially for low-resource languages, where LLMs can serve as a critical tool to break research bottlenecks, enhancing the understanding, analysis, and preservation of these languages. However, despite architectural innovations and expanded capabilities, fundamental limitations persist in adapting LLMs to low-resource languages.

\subsubsection{Gaps Between Available Corpora, Model Capabilities, and Research Needs}

\textbf{Availability of Corpora}
The available corpora for low-resource languages are diverse, but they are also fraught with issues. Sources include digitization projects of historical documents, language archives from specific regions, and limited academic research collections. These sources remain scattered and significantly smaller in scale compared to high-resource languages. While expanded context windows in models like GPT-5 (400K tokens) and Claude Sonnet 4 (1M tokens) can accommodate entire small language corpora~\cite{anthropic2025context}, this does not solve the fundamental problem of data quality. The lack of standardized curation processes leads to numerous annotation errors and irregular records, which even sophisticated architectures cannot fully compensate for. Furthermore, many low-resource languages exist primarily in oral traditions, and their digital presence remains minimal despite recent digitization efforts.

\textbf{Model Capabilities}
Large language models have evolved significantly in addressing low-resource language challenges, yet fundamental limitations persist. While GPT-5's unified reasoning architecture can dynamically adjust computational depth based on linguistic complexity~\cite{openai2025gpt5}, and Claude 4's extended thinking enables thorough analysis of limited linguistic data~\cite{anthropic2025claude4}, these advances primarily benefit languages with at least moderate digital presence. 

The introduction of MoE architectures allows for language-specific parameter allocation~\cite{zhou2022moe}, preventing interference between high and low-resource languages. However, the routing mechanisms themselves are trained predominantly on high-resource language patterns, potentially misallocating computational resources for truly low-resource languages. Chain-of-thought reasoning has shown impressive results—achieving 33.8\% performance gains with just 1,000 reasoning traces in languages like Swahili~\cite{yeo2025longcot}—but this assumes sufficient conceptual overlap between pivot and target languages, an assumption that fails for linguistically isolated languages.

Safe-completion frameworks in GPT-5 help prevent hallucination when operating at the edge of model knowledge~\cite{openai2025system}, yet this often results in models being overly cautious with low-resource languages, refusing to generate potentially useful outputs due to uncertainty. The models still struggle with unique grammatical structures, tonal systems, and non-concatenative morphology common in many low-resource languages.

\textbf{Research Needs}
Humanities research demands precise understanding of semantics and cultural connotations in ancient low-resource languages, comprehensive handling of minority language texts, and deep insights into historical language evolution. Current models, despite their advances, face specific challenges in meeting these needs:

\begin{enumerate}
    \item \textbf{Ancient Text Interpretation}
    \begin{itemize}
        \item Requires understanding not just language but also historical context and cultural symbolism
        \item Extended context windows can include relevant historical documents, but models lack deep cultural knowledge embedded in centuries of scholarship
        \item \textit{Example:} Classical Chinese texts often require understanding of historical allusions and philosophical concepts that cannot be inferred from text alone
    \end{itemize}
    
    \item \textbf{Minority Culture Preservation}
    \begin{itemize}
        \item Models must handle code-switching, dialectal variations, and culturally specific expressions
        \item Dynamic routing in GPT-5 can allocate resources for complex linguistic phenomena, but cannot compensate for the absence of cultural knowledge that native speakers possess implicitly
        \item Models struggle with complex metaphors, symbols, and culturally loaded content, creating a significant barrier
    \end{itemize}
    
    \item \textbf{Language Evolution Analysis}
    \begin{itemize}
        \item Researchers need models that can trace semantic shifts, grammatical changes, and lexical borrowings across time periods
        \item While models can process historical texts in sequence, they lack the theoretical linguistic framework to identify and explain evolutionary patterns systematically
        \item The gap between current model capabilities and specialized research needs remains substantial, particularly for languages without extensive scholarly documentation
    \end{itemize}
\end{enumerate}

\subsubsection{Special Requirements for Low-Resource Language Data and Limitations in Generative Technology}

\textbf{Special Requirements for Low-Resource Language Data}
From a data perspective, the scarcity of corpus data for low-resource languages is a critical factor limiting the further training of LLMs and their effective application in low-resource language research. The existing corpora, both in quantity and quality, are insufficient to meet the needs to improve LLM performance. To better utilize LLMs for low-resource language research, it is urgent to establish a specialized corpus for low-resource languages. This corpus should cover various aspects of low-resource languages, including different historical periods, regional variants, and text types. Additionally, high-quality data annotation is essential, with annotations including grammatical structures, semantic information, and cultural contexts, to improve data usability and value, providing better training materials for LLMs.

\textbf{Limitations in Generative Technology}
Current LLM technology has some inherent limitations when dealing with low-resource languages. The pre-training data primarily consists of large-scale text collections dominated by high-resource languages, causing the models to overlook low-resource languages during initial learning. Consequently, when processing related tasks, the models lack prior knowledge, such as the ability to capture unique vocabulary usage, grammatical rules, and semantic networks in low-resource languages. While MoE architectures allow for language-specific parameters, the routing mechanisms themselves are trained primarily on high-resource language patterns, potentially misallocating computational resources for truly low-resource languages. Chain-of-thought reasoning, though powerful, assumes sufficient conceptual overlap between pivot and target languages—an assumption that fails for linguistically isolated languages. Furthermore, the impressive context windows of modern models cannot substitute for the deep, implicit knowledge that emerges only through extensive pretraining on diverse linguistic data.


\section{Applications of Large Language Models in Linguistic Research}

In the context of low-resource languages (LRL), the value of LLMs lies not in one-time "high scores" or "demonstration effects," but in their potential to function as tools for linguistic research—facilitating data construction and hypothesis generation while maintaining reproducibility, interpretability, and community governance. Reducing LLMs to mere "task solvers" risks falling into the trap of metric fixation; in contrast, organizing research around the principle of "linguistic facts—theoretical propositions—evidence chains" enables models to serve as academic instruments rather than mere ranking machines. This chapter is structured around three central themes: first, role delineation—determining when LLMs should function as "pilots" and when as "co-pilots" within the research process; second, internal mechanisms and adaptation—how evidence from the multilingual "core" and language-specific "peripheries" can inform robust engineering and evaluation practices; third, space-time and typology—how credible discoveries and falsifiable claims can be established across the dimensions of geography, chronology, and linguistic typology. To illustrate these points, we integrate recent high-quality studies on diachronic semantics and public discourse into the relevant sections, demonstrating how "model capabilities" can be translated into "falsifiable linguistic knowledge."

\subsection{Large Language Models in Linguistic Research: the Pilot and the Copilot}

The distinction between ``pilot'' and ``copilot'' roles is not rhetorical flourish but a methodological commitment that determines how expertise and automation should be balanced in linguistic research. In the \textbf{copilot capacity}, the expert remains the primary decision-maker, while the model acts as an assistant, offering candidate analyses with supporting rationales, surfacing zones of uncertainty, and systematically feeding disagreements back into the dataset and annotation codebooks. The crucial point is that the human remains epistemically responsible: the model accelerates, but does not override, the processes of adjudication, theorization, and data governance.

Concrete examples make this distinction clearer. The participatory machine translation initiative across Africa, known as \textbf{Masakhane} \cite{orife2020masakhanemachinetranslation}, illustrates how collective engagement of local experts in sampling, labeling, and evaluation produces improvements on several dimensions simultaneously: higher annotation quality, greater reusability of data across projects, and increased social legitimacy of downstream applications. This is not simply a matter of accuracy. The collaborative loop fosters community trust and ensures that the resulting resources align with local priorities and linguistic realities---an essential criterion in domains where language is bound up with identity and sovereignty. A similar logic underpins documentation workflows such as \textbf{GlossLM} \cite{ginn2024glosslmmassivelymultilingualcorpus}, which leverages cross-lingual pretraining on interlinear glossed text to provide word-by-word suggestions across thousands of under-documented languages. The system may propose morpheme boundaries, glosses, or candidate syntactic structures, thereby accelerating tedious steps. Yet, the final quality is not determined by the system itself: it depends on human adjudication and on the continual refinement of annotation codebooks. In this sense, the workflow forms a closed feedback loop: copilot suggestion, expert adjudication, codebook feedback, improved copilot suggestions. What emerges is not a substitution of human expertise but a cyclical co-evolution of human conventions and machine affordances.

Then, when might an LLM legitimately assume the role of pilot? Importantly, this decision cannot rest on intuitive judgments about model fluency or perceived intelligence. Instead, it requires falsifiable criteria. The ``Alternative Annotator Test'' (AAT) \cite{calderon-etal-2025-alternative} offers such a criterion: a model may replace or precede human annotators only when (i) pre-established quality and cost thresholds are met within a well-defined domain, and (ii) the model’s failure modes are both interpretable and constrained. This test treats the model as one more annotator whose reliability must be benchmarked against inter-annotator agreement among humans \cite{artstein-poesio-2008-survey,Braylan_2022}. A model that passes AAT is not assumed to be perfect, but to function as an alternative annotator whose strengths and weaknesses are transparent, auditable, and integrated into governance protocols. 

This same rationale applies beyond annotation and into analytic domains such as \textbf{diachronic semantics} \cite{hamilton-etal-2016-diachronic,schlechtweg-etal-2020-semeval}. Consider workflows that combine LLM-driven word-sense induction (WSI) over temporal slices with regression analysis of sense proportions as functions of time, speaker age, and institutional role \cite{amrami2019bettersubstitutionbasedwordsense,Giulianelli_2020,dicarlo2019trainingtemporalwordembeddings}. The copilot mode allows models to generate candidate sense clusters, align them across historical periods, and surface statistical discrepancies, while human reviewers validate or challenge these outputs. Such a division of labor makes possible new scales of inquiry while preserving scholarly accountability. The point is not to outsource interpretation but to reconfigure it: the model becomes a hypothesis generator, while the human expert adjudicates theoretical import. Evidence from large-scale studies reinforces this point. A recent paper in \emph{PNAS} \cite{doi:10.1073/pnas.2426815122} analyzing 7.9 million U.S. Congressional speeches (1873--2010) demonstrated that semantic change is not primarily generational. Instead, speakers across all age groups adapt to contemporary semantic trends, with older speakers typically lagging behind by only two to three years and, in some cases, even leading sense innovation \cite{kamath.2025.semantic}. Such findings underscore the necessity of rigorous, large-scale methods and also highlight why interpretability and transparency in LLM-driven pipelines are indispensable: without careful human-guided validation, one risks mistaking spurious statistical shifts for genuine semantic change.

In summary, the pilot/copilot metaphor provides a principled framework for calibrating the balance of human and machine roles in linguistic research. In copilot mode, models accelerate, suggest, and highlight; in pilot mode, they may, under strict falsifiable conditions, provisionally replace human annotators. But in both roles, success depends on robust feedback loops, transparent governance, and sustained expert oversight. Only then can LLMs become not just tools of efficiency but instruments of reliable, socially accountable knowledge production.

\subsection{Unveiling Linguistic Regions in LLMs}

\subsubsection{Automatic recognition and modeling of language variants and dialect differences}

The automatic recognition of dialects and language variants has become a central challenge for multilingual NLP. Traditional language identification (LID) models often treat ``language'' as a single, globally uniform category, overlooking the fact that intra-language variation can be as great as or greater than inter-language differences. To address this, it is essential to treat both \emph{geographic space} and \emph{variant mixture} as first-class signals, rather than as noise to be suppressed.  

\textbf{GeoAdapt}, a recent approach \cite{hofmann-etal-2024-geographic}, demonstrates the effectiveness of injecting \emph{geolinguistic signals} directly into encoder architectures. Instead of relying on a single global classifier, which risks overgeneralization and false positives, GeoAdapt leverages region-conditioned plausible language sets \cite{wang2025geolocationawarerobustspokenlanguage}. This strategy narrows the hypothesis space to dialects actually spoken in a given region, significantly reducing misclassification rates at scale. For example, in multilingual contexts such as Switzerland, studies on Swiss German reveal that while there is a robust and genuine dialectal signal detectable in embeddings, the proximity of closely related varieties leads to persistent confusion \cite{vamvas-etal-2024-modular}. This highlights an important methodological pitfall: reporting only macro-averaged accuracy across languages obscures dialect-level failures and paints an overly optimistic picture of performance.  

Dialect-sensitive modeling is equally crucial in settings involving \emph{code-switching}. While LLMs now consistently outperform strong supervised baselines on benchmarked code-switching tasks \cite{khanuja-etal-2020-gluecos}, their performance deteriorates when confronted with more fluid, naturally occurring mixing patterns or when faced with domain shifts such as informal social media text. This brittleness underscores the need for stress-testing models under realistic multilingual conditions, rather than relying solely on curated datasets.  

In the Arabic NLP community, the introduction of \textbf{AraDiCE} provides a concrete response to these challenges. Unlike many prior resources that implicitly privilege Modern Standard Arabic (MSA), AraDiCE explicitly pairs dialectal coverage with cultural knowledge \cite{mousi-etal-2025-aradice}. This design avoids ``standard-language centrism,'' ensuring that dialects are treated as primary linguistic objects rather than deviations. Moreover, AraDiCE introduces quantifiable measures of coverage across dialect regions, allowing researchers to more transparently assess which varieties are well supported and which remain underrepresented \cite{mousi-etal-2025-aradice}.  

\textit{Practice rules.}  
Building on these insights, several best practices can be articulated for dialect-aware modeling:  
\begin{itemize}
    \item \textbf{Audit and control for toponym shortcuts:} models may learn to rely on explicit place names (``Cairo,'' ``Zurich'') rather than genuine linguistic cues. To mitigate this, it is preferable to use coarse regional metadata, and when necessary, remove place names altogether during training and evaluation \cite{hofmann-etal-2024-geographic}.    
    \item \textbf{Incorporate code-switch stress testing by default:} model evaluation should include tasks with rapid and fluid switching, reflecting authentic multilingual discourse rather than only balanced, neatly segmented datasets \cite{khanuja-etal-2020-gluecos}.  
    \item \textbf{Stratify metrics within a language by variety or region:} rather than reporting only cross-language means, which may obscure intra-language disparities, fine-grained stratification reveals performance trade-offs and highlights which dialects remain underserved \cite{faisal-etal-24-dialectbenchy}.  
\end{itemize}

Taken together, these practices push the field toward a more equitable and robust treatment of linguistic diversity. Automatic recognition and modeling of language variants is not only a technical challenge but also a sociolinguistic imperative, ensuring that NLP systems reflect the full spectrum of human language use rather than a narrow, standardized subset.  

\subsubsection{Internal mechanisms and cross-language generalization: a multilingual core and monolingual regions}
\label{sec:multilingual-core-mono-regions}

Multilingual Transformers exhibit a \emph{sparse, modular} organization: a compact \textbf{multilingual core} carries language-agnostic structure and meaning, while \textbf{language-/family-specific regions} capture orthography, morphology, and idiosyncratic constructions \cite{foroutan-etal-2022-discovering,pires-etal-2019-multilingual,tang-etal-2024-language}. This motivates \emph{freeze-and-specialize}: keep the core stable and attach small per-language adapters/LoRA or sparsely routed experts to absorb variation, reducing cross-language interference under fixed compute and mitigating the “curse of multilinguality” \cite{pfeiffer-etal-2020-mad,shazeer2017moe,fedus2022switch,arivazhagan2019massively}.

\paragraph{Conceptual split.}
Two principles guide clean separation: (i) \textbf{script invariance} in the core (surface differences map to similar internal states); (ii) \textbf{decorrelation} so language-specific factors reside in local modules. A practical diagnostic is reduced \textbf{gradient cross-talk} in the core after specialization.

\paragraph{Evidence (condensed).}
\begin{itemize}
  \item \textbf{Representations:} cross-script pairs align in core layers; clustering follows syntax/semantics rather than script \cite{pires-etal-2019-multilingual,chang2022geometry}.
  \item \textbf{Causality:} disabling locals hurts orthography/morphology with minor effect on cross-lingual semantics; core perturbations show the converse \cite{pfeiffer-etal-2020-mad,foroutan-etal-2022-discovering}.
  \item \textbf{Routing/sparsity:} MoE gates differentiate languages; localized circuits emerge for language-specific phenomena \cite{shazeer2017moe,fedus2022switch}.
  \item \textbf{Dynamics:} full finetuning forgets more; frozen cores + small locals yield a better stability–specialization trade-off \cite{winata-etal-2023-overcoming}.
\end{itemize}

\paragraph{Protocol (actionable).}
\begin{enumerate}
  \item \textbf{Core pretrain then freeze:} broad multilingual data, paired transliteration and script perturbations, morphology process probes.
  \item \textbf{Localize:} fit small per-language \& family modules; for MoE use top-$k$ routing and gate-entropy penalties; for adapters \& LoRA enforce sparsity \& low rank \cite{shazeer2017moe,fedus2022switch,pfeiffer-etal-2020-mad}.
  \item \textbf{Consolidate (optional):} distill core and locals into a sparse expert layer to retain modularity at low latency \cite{zhang-etal-2024-lightweight}.
\end{enumerate}

\paragraph{LRL first principles.}
\textbf{Script \& morphology engineering} often dominates scale: (i) reversible, traceable normalization \& transliteration with provenance \cite{xhelili-etal-2024-breaking}; (ii) deliberate tokenization (byte \& char \& subword) by script and typology \cite{yu2020pcgrad}; (iii) segmentation \& reinflection as process-level checks so gains reflect structure, not orthography \cite{shazeer2017moe,fedus2022switch,winata-etal-2023-overcoming}.

\paragraph{Evaluation \& diagnostics.}
Report both \emph{task} and \emph{mechanistic} signals: cross-script alignment and round-trip transliteration accuracy; gradient cross-talk before \& after specialization; routing statistics (by language and construction); causal ablations (zero gates \& mask locals \& activation patching); and forgetting curves when adding new languages.

\paragraph{Threats \& practice (brief).}
Guard against transliteration leakage, domain–language confounds, and probe overfitting; co-report mechanistic indicators alongside task scores. Release preprocessing code and logs, parameter \& routing settings, and reproducible scripts for diagnostics.

\subsection{Opportunities and challenges: from making results to making institutions}
\textbf{Opportunities.} (i) Rewiring the division of labor: copilot workflows shift expert time from repetitive tasks to theory and design, leaving durable data and norms; (ii) Closing the mechanism$\rightarrow$method loop: the core \& region picture operationalizes as freeze+specialize or expert routing, yielding more stable coverage at fixed compute; (iii) Bringing space, time, and phenomena into the \emph{main} evaluation layers (WSI-level semantics, region-conditioned LID, code-switch stress), so \emph{linguistic facts} rather than frequency artifacts drive conclusions.

\textbf{Challenges.} (a) Structural inequality---without stratification and audits, global means can mask regress for marginal languages; (b) Contamination and leakage---in small LRL testbeds, tiny leaks produce fake wins, so \emph{leakage visibility} must be standard; (c) Energy and governance costs---to be reported alongside performance. The answer is not retreating to Anglocentric convenience but institutional commitments: participatory sampling and labeling; mandatory governance docs; reproducible contamination audits; energy/carbon co-reporting; and inclusion of code-switching and multimodality in \emph{mainline} evaluation.

When copilot is the default, governance the prior, T$\times$G$\times$T the scaffold, the core/region mechanism operationalized (freeze+specialize), and diachronic semantics plus code-switching become \emph{standard} evaluation layers, LLMs shift from “doing tasks” to “producing knowledge” for LRLs. The path is already visible: Masakhane’s participatory MT; GlossLM for cross-lingual documentation; GeoAdapt \& GeoLID for spatial signals; AraDiCE for culture$\leftrightarrow$dialect; temporal alignment and era-restricted pretraining; WSI with age \& role regression; and expert \& modular multilinguality for coverage and robustness. Next, run the same pipeline across typologically distinct families (e.g., Bantu, Arabic dialects, Dravidian) to turn “low resource” into a \emph{high-value source of linguistic knowledge}.

\section{Future Research Directions}

\subsection{Development of Specialized Language Models for Low-Resource Languages}

\begin{itemize}
    \item \textbf{Data Augmentation Techniques for Limited Corpora}

    Data augmentation techniques can significantly mitigate the issue of data scarcity in low-resource languages. Methods such as back-translation, synthetic data generation, and contextual data augmentation have been shown to increase the volume and diversity of training data. For instance, back-translation can use high-resource languages as intermediaries to generate pseudo-parallel corpora, while techniques like paraphrasing or noise injection introduce linguistic variability. These approaches ensure better model generalization and capture unique language-specific patterns that would otherwise remain underrepresented.

    \item \textbf{Multilingual Pretraining for Cross-Language Knowledge Transfer}

    Leveraging multilingual pretraining allows large language models to benefit from shared linguistic features across languages. Future research could enhance this approach by exploring dynamic pretraining strategies, where the model progressively adjusts its focus to prioritize low-resource languages during training. Incorporating typological features can further improve transferability, especially for languages with rich morphological or syntactic characteristics.

    \item \textbf{Few-Shot and Zero-Shot Learning for Rapid Adaptation}

    Few-shot and zero-shot learning provide avenues for adapting models to low-resource languages with minimal labeled data. Beyond common approaches like prompt-based learning, innovative frameworks such as Model-Agnostic Meta-Learning (MAML)~\cite{finn2017model} optimize models for task adaptability, enabling them to perform well with only a handful of examples. Dynamic task re-weighting can prioritize features unique to low-resource languages, ensuring better alignment with linguistic structures. Additionally, task-conditioning embeddings that incorporate linguistic typology can guide models to focus on crucial language-specific attributes. For zero-shot learning, retrieval-augmented reasoning can dynamically fetch relevant multilingual examples, improving the model's context-aware adaptation to unseen languages.

    \item \textbf{Incorporating Sociolinguistic and Dialectal Variations}

    Low-resource languages often exhibit substantial dialectal and sociolinguistic diversity, which poses challenges for language modeling. To address this, researchers can design models with dialectal embeddings that capture regional and social variations. Another approach is to incorporate multi-variant corpora that represent a broad spectrum of dialects, ensuring that the model generalizes effectively across linguistic communities. Robust pretraining strategies that integrate both written and spoken forms of a language can help models better reflect real-world usage, making them more adaptable to diverse linguistic contexts.

\end{itemize}

\subsection{Interdisciplinary Collaboration for Low-Resource Language Preservation}

\begin{itemize}
    \item \textbf{Collaborations Between Linguists and AI Researchers}

    Effective preservation of low-resource languages requires collaboration between linguists and AI researchers to bridge the gap between linguistic theory and computational implementation. Linguists can provide insights into unique grammatical structures, phonetic systems, and language-specific phenomena, while AI researchers design models that can effectively leverage these features. For example, linguist-informed annotated corpora can guide language models to better understand syntax and morphology, improving translation, transcription, and other NLP tasks. This partnership ensures that AI solutions respect and reflect the linguistic diversity and cultural depth of low-resource languages.

    \item \textbf{Integration of Anthropological and Ethnographic Perspectives}

    Anthropologists and ethnographers bring invaluable cultural and societal context to language preservation efforts. Their expertise in documenting oral traditions, cultural practices, and regional nuances can enrich AI models with contextually relevant data. For instance, ethnographic insights can help define the sociolinguistic parameters of a language, such as its use in ceremonies or as a marker of identity within a community. By combining this knowledge with AI's data-processing capabilities, interdisciplinary projects can ensure that the cultural essence of a language is preserved alongside its linguistic form.

    \item \textbf{Role of Philology and Historical Studies}

    Philologists and historians play a crucial role in digitizing and contextualizing ancient texts in low-resource languages. Their expertise in deciphering scripts, annotating ancient documents, and reconstructing linguistic evolution complements AI's ability to process large datasets and generate insights. Collaborative efforts can focus on creating searchable archives of ancient manuscripts, improving access to historical resources, and training AI models to interpret historical language variants. This ensures that both contemporary and historical dimensions of low-resource languages are preserved.

    \item \textbf{Community-Centric Data Collection and Annotation}

    Engaging native speakers and local communities in data collection and annotation ensures that the linguistic and cultural authenticity of low-resource languages is maintained. Community-driven initiatives, such as crowd-sourcing linguistic data or organizing language documentation workshops, empower speakers to take ownership of preservation efforts. By incorporating their perspectives, interdisciplinary projects can build more comprehensive and representative datasets, while fostering a sense of cultural pride and involvement in the digital preservation process.

    \item \textbf{Educational Tools and Language Revitalization Programs}

    Interdisciplinary collaboration can also focus on creating educational tools that support language revitalization. For example, linguists, educators, and technologists can co-develop language learning apps, interactive dictionaries, and digital games tailored to teaching low-resource languages. These tools not only help younger generations reconnect with their linguistic heritage but also ensure that the language remains active and evolving. Combining pedagogy with AI-driven insights enables the development of engaging and effective resources that address the specific needs of diverse learner groups.

\end{itemize}

\subsection{Potential for Social Innovation and Cultural Dissemination Through Low-Resource Language Models}

\begin{itemize}
    \item \textbf{Preserving Endangered Languages Through Digital Tools}

    Low-resource language models can serve as powerful tools for documenting and preserving endangered languages. By creating accessible and interactive digital platforms, these models enable the recording of oral traditions, folklore, and cultural narratives, ensuring their longevity. For example, speech-to-text systems powered by language models can transcribe endangered languages in real time, while mobile applications can gamify language learning to attract younger audiences. These efforts help maintain the cultural identity of minority communities and foster intergenerational transmission of linguistic heritage.

    \item \textbf{Facilitating Cross-Cultural Communication and Collaboration}

    Low-resource language models can bridge linguistic gaps between communities, fostering greater cross-cultural understanding and collaboration. For instance, real-time translation tools can enable seamless communication in multilingual regions, promoting social integration and reducing linguistic barriers. Additionally, cultural exchanges facilitated by these technologies can strengthen connections between communities, enabling them to share their histories, traditions, and values on a global stage.

    \item \textbf{Empowering Communities with Accessible Education and Resources}

    Language models tailored to low-resource languages can play a transformative role in education. By providing interactive learning tools, digital libraries, and culturally relevant educational content, these models make language resources accessible to marginalized communities. For example, AI-powered chatbots can support language learners with personalized feedback, while digital storytelling platforms can deliver culturally specific narratives. These innovations not only preserve linguistic diversity but also promote equity by ensuring that underrepresented communities have access to modern educational opportunities.

    \item \textbf{Revitalizing Cultural Practices and Artistic Traditions}

    By digitizing and analyzing cultural expressions such as poetry, songs, and oral storytelling, low-resource language models can aid in the revival of traditional art forms. AI tools can assist in creating modern adaptations of these traditions, such as composing music inspired by folk songs or generating digital artwork based on traditional motifs. These efforts not only preserve cultural practices but also make them more accessible and appealing to contemporary audiences, encouraging their continued relevance and evolution.

    \item \textbf{Driving Economic and Social Development in Minority Communities}

    By integrating low-resource languages into digital and business ecosystems, language models can drive economic growth and social empowerment. For instance, localized AI applications, such as customer service chatbots or e-commerce platforms, can cater to minority language speakers, opening up new markets and opportunities. Additionally, language models can support community-based tourism initiatives by translating cultural information and providing accessible guides in native languages, boosting local economies while promoting cultural heritage.

    \item \textbf{Ensuring Ethical and Inclusive Cultural Dissemination}

    While low-resource language models offer significant potential for cultural dissemination, ethical considerations must remain central to their design and application. Developers must engage with native speakers and community leaders to ensure that cultural knowledge is represented accurately and shared with consent. Transparent data practices, fair compensation for contributors, and respect for intellectual property rights are critical for fostering trust and preventing exploitation. Inclusive frameworks can ensure that technological advancements benefit the communities they aim to serve.
    
\end{itemize}

\section{Conclusion}

The study of low-resource languages, despite their cultural, historical, and intellectual significance, remains an underserved area in computational linguistics. These languages, as vital repositories of humanity’s diverse heritage, face the dual threat of underrepresentation in digital ecosystems and extinction in the real world. This paper has explored how large language models can serve as transformative tools for preserving and revitalizing these languages, offering new avenues for research, cultural preservation, and cross-disciplinary collaboration.

LLMs, with their capacity for multilingual understanding and contextual adaptation, have demonstrated immense potential to address the unique challenges posed by low-resource languages. From breakthroughs in data augmentation and cross-language transfer to innovative methods like zero-shot learning and dialect embedding, these models provide scalable solutions to bridge linguistic gaps. By leveraging techniques such as retrieval-augmented generation, task-conditioning embeddings, and meta-learning, LLMs are increasingly capable of adapting to the intricate linguistic and sociocultural dimensions of these languages, fostering their inclusion in the global linguistic landscape.

However, this journey is fraught with challenges. The scarcity and complexity of data, coupled with the computational and ethical considerations inherent in working with marginalized languages, underscore the need for sustained, interdisciplinary efforts. Linguists, anthropologists, educators, and AI researchers must collaborate to ensure that the technological advancements are culturally sensitive, ethically grounded, and community-driven. Efforts to build annotated corpora, digitize historical texts, and engage native speakers in the data collection process are not merely technical necessities but ethical imperatives.

Furthermore, the application of low-resource language models transcends linguistic preservation, opening new pathways for social innovation and economic empowerment. By fostering cross-cultural communication, enhancing education accessibility, and revitalizing artistic traditions, these models hold the promise of creating a more inclusive and interconnected world. At the same time, careful attention must be given to ethical data practices, ensuring transparency, fairness, and respect for the intellectual property and cultural sovereignty of indigenous and minority communities.

In conclusion, the convergence of advanced AI capabilities and the collective commitment of global communities offers an unprecedented opportunity to safeguard humanity’s linguistic and cultural diversity. Low-resource language models are not merely tools for computational innovation but catalysts for preserving the narratives, knowledge systems, and identities that define our shared heritage. By prioritizing inclusivity, ethical engagement, and interdisciplinary collaboration, we can ensure that these languages thrive in both digital and social realms, contributing to a richer and more equitable future for all.


\bibliographystyle{splncs04}
\bibliography{mybib} 

@inbook{Sacred_Texts_Digital_Age,
author = {Anderson, Bradford},
year = {2020},
month = {06},
pages = {281-302},
title = {Sacred Texts in a Digital Age: Materiality, Digital Culture, and the Functional Dimensions of Scriptures in Judaism, Christianity, and Islam},
isbn = {9783110634440},
publisher = "De Gruyter",
doi = {10.1515/9783110634440-013}
}

@inproceedings{lai-2024-tupleised,
    title = "Tupleised co-occurrence measures vs {LLM} word embeddings for corpus linguistics: The case of {E}nglish light verb construction detection",
    author = "Lai, Ryan Ka Yau",
    editor = "Oco, Nathaniel  and
      Dita, Shirley N.  and
      Borlongan, Ariane Macalinga  and
      Kim, Jong-Bok",
    booktitle = "Proceedings of the 38th Pacific Asia Conference on Language, Information and Computation",
    month = dec,
    year = "2024",
    address = "Tokyo, Japan",
    publisher = "Tokyo University of Foreign Studies",
    url = "https://aclanthology.org/2024.paclic-1.116/",
    pages = "1201--1212"
}

@article{Mahowald_2024,
author = {Mahowald, Kyle and Ivanova, Anna and Blank, Idan and Kanwisher, Nancy and Tenenbaum, Joshua and Fedorenko, Evelina},
year = {2024},
month = {03},
pages = {},
title = {Dissociating language and thought in large language models},
volume = {28},
journal = {Trends in Cognitive Sciences},
doi = {10.1016/j.tics.2024.01.011}
}

@article{Medine_2022, title={bell hooks, Black Feminist Thought, and Black Buddhism: A Tribute}, volume={7}, url={https://scholarworks.iu.edu/iupjournals/index.php/jwp/article/view/5479}, abstractNote={&amp;lt;p&amp;gt;This tribute to the late bell hooks examines her work as a Black feminist and Black Buddhist. After a brief introduction to her life, I examine her contributions to feminist thought, particularly her understanding of the need to dismantle “imperial white supremacist capitalist patriarchy.” As a Black feminist and woman, hooks comes to this work, first, with rage, but in her turn to Buddhist thought, she develops a love ethic, one that she wrote extensively about until her death in 2021 of renal failure.&amp;lt;/p&amp;gt;}, number={1}, journal={Journal of World Philosophies}, author={Medine, Carolyn M. Jones}, year={2022}, month={Jul.}, pages={187–196} }

@inproceedings{lai-etal-2023-turn,
    title = "Turn design, resonance and epistemic stance in the Diamond Sutra: A dialogic constructionist approach",
    author = "Lai, Ryan Ka Yau  and
      Yin, Lily Zihe  and
      Zhang, Alice Yimeng  and
      Jiang, Yuting  and
      Xin, Bill Shiyang  and
      Gao, Junwei",
    editor = "Huang, Chu-Ren  and
      Harada, Yasunari  and
      Kim, Jong-Bok  and
      Chen, Si  and
      Hsu, Yu-Yin  and
      Chersoni, Emmanuele  and
      A, Pranav  and
      Zeng, Winnie Huiheng  and
      Peng, Bo  and
      Li, Yuxi  and
      Li, Junlin",
    booktitle = "Proceedings of the 37th Pacific Asia Conference on Language, Information and Computation",
    month = dec,
    year = "2023",
    address = "Hong Kong, China",
    publisher = "Association for Computational Linguistics",
    url = "https://aclanthology.org/2023.paclic-1.75/",
    pages = "753--763"
}

@inproceedings{vasconcelos2024disappearing,
  title={Disappearing without a Trace: Coverage, Community, Quality, and Temporal Dynamics of Wikipedia Articles on Endangered Brazilian Indigenous Languages},
  author={Vasconcelos, Marisa and de Souza Mizukami, Priscila and Pinhanez, Claudio Santos},
  booktitle={Proceedings of the International AAAI Conference on Web and Social Media},
  volume={18},
  pages={1531--1544},
  year={2024}
}

@article{dai2023chataug,
  title={Chataug: Leveraging chatgpt for text data augmentation},
  author={Dai, Haixing and Liu, Zhengliang and Liao, Wenxiong and Huang, Xiaoke and Wu, Zihao and Zhao, Lin and Liu, Wei and Liu, Ninghao and Li, Sheng and Zhu, Dajiang and others},
  journal={arXiv preprint arXiv:2302.13007},
  year={2023}
}

@article{tian2024assessing,
  title={Assessing Large Language Models in Mechanical Engineering Education: A Study on Mechanics-Focused Conceptual Understanding},
  author={Tian, Jie and Hou, Jixin and Wu, Zihao and Shu, Peng and Liu, Zhengliang and Xiang, Yujie and Gu, Beikang and Filla, Nicholas and Li, Yiwei and Liu, Ning and others},
  journal={arXiv preprint arXiv:2401.12983},
  year={2024}
}

@article{huang2024trustllm,
  title={Trustllm: Trustworthiness in large language models},
  author={Huang, Yue and Sun, Lichao and Wang, Haoran and Wu, Siyuan and Zhang, Qihui and Li, Yuan and Gao, Chujie and Huang, Yixin and Lyu, Wenhan and Zhang, Yixuan and others},
  journal={arXiv preprint arXiv:2401.05561},
  year={2024}
}

@article{openai2025gpt5,
  title={Introducing gpt-5,},
  author={OpenAI},
  year={2025},
  month={August},
  journal={openai},
  url={https://openai.com/gpt-5/}
}

@article{google2025gemini,
  title={Gemini 2.5: Our most intelligent AI model},
  author={Google DeepMind},
  year={2025},
  month={March},
  journal={google},
  url={https://blog.google/technology/google-deepmind/gemini-model-thinking-updates-march-2025/#gemini-2-5-thinking}
}

@article{anthropic2025claude4,
  title={Introducing Claude 4},
  author={Anthropic},
  year={2025},
  month={May},
  journal={anthropic},
  url={https://www.anthropic.com/news/claude-4}
}

@article{stanford2025hai,
  title={Mind the (Language) Gap},
  author={Pava, Juan N and Meinhardt, Caroline and Zaman, Haifa Badi Uz and Friedman, Toni and Truong, Sang T and Zhang, Daniel and Marivate, Vukosi and Koyejo, Sanmi},
  journal={stanfordhai},
  year={2025}
}

@article{chen2024queen,
  title={Queen: A large language model for quechua-english translation},
  author={Chen, Junhao and Shu, Peng and Li, Yiwei and Zhao, Huaqin and Jiang, Hanqi and Pan, Yi and Zhou, Yifan and Liu, Zhengliang and Howe, Lewis C and Liu, Tianming},
  journal={arXiv preprint arXiv:2412.05184},
  year={2024}
}

@article{gao2025tlue,
  title={TLUE: A Tibetan language understanding evaluation benchmark},
  author={Gao, Fan and Huang, Cheng and Tashi, Nyima and Wang, Xiangxiang and Tsering, Thupten and Ma-bao, Ban and Duojie, Renzeg and Luosang, Gadeng and Dongrub, Rinchen and Tashi, Dorje and others},
  journal={arXiv preprint arXiv:2503.12051},
  year={2025}
}

@inproceedings{ebrahimi-2022-americasnli,
    title = "{A}mericas{NLI}: Evaluating Zero-shot Natural Language Understanding of Pretrained Multilingual Models in Truly Low-resource Languages",
    author = "Ebrahimi, Abteen  and
      Mager, Manuel  and
      Oncevay, Arturo  and
      Chaudhary, Vishrav  and
      Chiruzzo, Luis  and
      Fan, Angela  and
      Ortega, John  and
      Ramos, Ricardo  and
      Rios, Annette  and
      Meza Ruiz, Ivan Vladimir  and
      Gim{\'e}nez-Lugo, Gustavo  and
      Mager, Elisabeth  and
      Neubig, Graham  and
      Palmer, Alexis  and
      Coto-Solano, Rolando  and
      Vu, Thang  and
      Kann, Katharina",
    booktitle = "Proceedings of the 60th Annual Meeting of the Association for Computational Linguistics (Volume 1: Long Papers)",
    month = may,
    year = "2022",
    address = "Dublin, Ireland",
    publisher = "Association for Computational Linguistics",
    url = "https://aclanthology.org/2022.acl-long.435",
    pages = "6279--6299",
}

@article{zhang2025milic,
  title={MiLiC-Eval: Benchmarking Multilingual LLMs for China's Minority Languages},
  author={Zhang, Chen and Tao, Mingxu and Liao, Zhiyuan and Feng, Yansong},
  journal={arXiv preprint arXiv:2503.01150},
  year={2025}
}

@inproceedings{zhang2022cherokee,
    title = "How can {NLP} Help Revitalize Endangered Languages? A Case Study and Roadmap for the {C}herokee Language",
    author = "Zhang, Shiyue  and
      Frey, Ben  and
      Bansal, Mohit",
    editor = "Muresan, Smaranda  and
      Nakov, Preslav  and
      Villavicencio, Aline",
    booktitle = "Proceedings of the 60th Annual Meeting of the Association for Computational Linguistics (Volume 1: Long Papers)",
    month = may,
    year = "2022",
    address = "Dublin, Ireland",
    publisher = "Association for Computational Linguistics",
    url = "https://aclanthology.org/2022.acl-long.108/",
    doi = "10.18653/v1/2022.acl-long.108",
    pages = "1529--1541"
}

@article{yeo2025longcot,
  title={Demystifying long chain-of-thought reasoning in llms},
  author={Yeo, Edward and Tong, Yuxuan and Niu, Morry and Neubig, Graham and Yue, Xiang},
  journal={arXiv preprint arXiv:2502.03373},
  year={2025}
}

@article{zhou2022moe,
  title={Mixture-of-experts with expert choice routing},
  author={Zhou, Yanqi and Lei, Tao and Liu, Hanxiao and Du, Nan and Huang, Yanping and Zhao, Vincent and Dai, Andrew M and Le, Quoc V and Laudon, James and others},
  journal={Advances in Neural Information Processing Systems},
  volume={35},
  pages={7103--7114},
  year={2022}
}

@misc{anthropic2025context,
  title={Claude Sonnet 4 now supports 1M tokens of context},
  author={{Anthropic}},
  year={2025},
  month={August},
  day={12},
  howpublished={\url{https://www.anthropic.com/news/1m-context}},
  note={Anthropic API Update}
}

@misc{openai2025system,
  title={GPT-5 System Card},
  author={{OpenAI}},
  year={2025},
  month={August},
  howpublished={\url{https://openai.com/index/gpt-5-system-card/}},
  note={OpenAI Safety Publication}
}

@article{liu2024surviving,
  title={Surviving ChatGPT in healthcare},
  author={Liu, Zhengliang and Zhang, Lu and Wu, Zihao and Yu, Xiaowei and Cao, Chao and Dai, Haixing and Liu, Ninghao and Liu, Jun and Liu, Wei and Li, Quanzheng and others},
  journal={Frontiers in Radiology},
  volume={3},
  pages={1224682},
  year={2024},
  publisher={Frontiers Media SA}
}

@inproceedings{huang2024position,
  title={Position: Trustllm: Trustworthiness in large language models},
  author={Huang, Yue and Sun, Lichao and Wang, Haoran and Wu, Siyuan and Zhang, Qihui and Li, Yuan and Gao, Chujie and Huang, Yixin and Lyu, Wenhan and Zhang, Yixuan and others},
  booktitle={International Conference on Machine Learning},
  pages={20166--20270},
  year={2024},
  organization={PMLR}
}

@article{liu2023radiology,
  title={Radiology-llama2: Best-in-class large language model for radiology},
  author={Liu, Zhengliang and Li, Yiwei and Shu, Peng and Zhong, Aoxiao and Yang, Longtao and Ju, Chao and Wu, Zihao and Ma, Chong and Luo, Jie and Chen, Cheng and others},
  journal={arXiv preprint arXiv:2309.06419},
  year={2023}
}

@article{lyu2024gp,
  title={GP-GPT: Large Language Model for Gene-Phenotype Mapping},
  author={Lyu, Yanjun and Wu, Zihao and Zhang, Lu and Zhang, Jing and Li, Yiwei and Ruan, Wei and Liu, Zhengliang and Yu, Xiaowei and Cao, Chao and Chen, Tong and others},
  journal={arXiv preprint arXiv:2409.09825},
  year={2024}
}

@article{wang2024comprehensive,
  title={A comprehensive review of multimodal large language models: Performance and challenges across different tasks},
  author={Wang, Jiaqi and Jiang, Hanqi and Liu, Yiheng and Ma, Chong and Zhang, Xu and Pan, Yi and Liu, Mengyuan and Gu, Peiran and Xia, Sichen and Li, Wenjun and others},
  journal={arXiv preprint arXiv:2408.01319},
  year={2024}
}

@article{liao2023differentiating,
  title={Differentiating ChatGPT-generated and human-written medical texts: quantitative study},
  author={Liao, Wenxiong and Liu, Zhengliang and Dai, Haixing and Xu, Shaochen and Wu, Zihao and Zhang, Yiyang and Huang, Xiaoke and Zhu, Dajiang and Cai, Hongmin and Li, Quanzheng and others},
  journal={JMIR Medical Education},
  volume={9},
  number={1},
  pages={e48904},
  year={2023},
  publisher={JMIR Publications Inc., Toronto, Canada}
}

@article{liu2023context,
  title={Context Matters: A Strategy to Pre-train Language Model for Science Education},
  author={Liu, Zhengliang and He, Xinyu and Liu, Lei and Liu, Tianming and Zhai, Xiaoming},
  journal={arXiv preprint arXiv:2301.12031},
  year={2023}
}

@inproceedings{rezayi2022clinicalradiobert,
  title={ClinicalRadioBERT: Knowledge-Infused Few Shot Learning for Clinical Notes Named Entity Recognition},
  author={Rezayi, Saed and Dai, Haixing and Liu, Zhengliang and Wu, Zihao and Hebbar, Akarsh and Burns, Andrew H and Zhao, Lin and Zhu, Dajiang and Li, Quanzheng and Liu, Wei and others},
  booktitle={Machine Learning in Medical Imaging: 13th International Workshop, MLMI 2022, Held in Conjunction with MICCAI 2022, Singapore, September 18, 2022, Proceedings},
  pages={269--278},
  year={2022},
  organization={Springer}
}

@article{dai2023ad,
  title={AD-AutoGPT: An Autonomous GPT for Alzheimer's Disease Infodemiology},
  author={Dai, Haixing and Li, Yiwei and Liu, Zhengliang and Zhao, Lin and Wu, Zihao and Song, Suhang and Shen, Ye and Zhu, Dajiang and Li, Xiang and Li, Sheng and others},
  journal={arXiv preprint arXiv:2306.10095},
  year={2023}
}

@article{zhao2023ophtha,
  title={Ophtha-llama2: A large language model for ophthalmology},
  author={Zhao, Huan and Ling, Qian and Pan, Yi and Zhong, Tianyang and Hu, Jin-Yu and Yao, Junjie and Xiao, Fengqian and Xiao, Zhenxiang and Zhang, Yutong and Xu, San-Hua and others},
  journal={arXiv preprint arXiv:2312.04906},
  year={2023}
}

@article{zhang2024generalist,
  title={A generalist vision--language foundation model for diverse biomedical tasks},
  author={Zhang, Kai and Zhou, Rong and Adhikarla, Eashan and Yan, Zhiling and Liu, Yixin and Yu, Jun and Liu, Zhengliang and Chen, Xun and Davison, Brian D and Ren, Hui and others},
  journal={Nature Medicine},
  pages={1--13},
  year={2024},
  publisher={Nature Publishing Group US New York}
}

@article{liu2023radonc,
  title={Radonc-gpt: A large language model for radiation oncology},
  author={Liu, Zhengliang and Wang, Peilong and Li, Yiwei and Holmes, Jason and Shu, Peng and Zhang, Lian and Liu, Chenbin and Liu, Ninghao and Zhu, Dajiang and Li, Xiang and others},
  journal={arXiv preprint arXiv:2309.10160},
  year={2023}
}

@article{liu2024fine,
  title={FINE-TUNING LARGE LANGUAGE MODELS FOR RADIATION ONCOLOGY, A HIGHLY SPECIALIZED HEALTHCARE DOMAIN},
  author={Liu, Zhengliang and Wang, Peilong and Li, Yiwei and Holmes, Jason M and Shu, Peng and Zhang, Lian and Li, Xiang and Li, Quanzheng and Vora, Sujay A and Patel, Samir and others},
  journal={International Journal of Particle Therapy},
  volume={12},
  pages={100428},
  year={2024},
  publisher={Elsevier}
}

@article{zhao2024revolutionizing,
  title={Revolutionizing finance with llms: An overview of applications and insights},
  author={Zhao, Huaqin and Liu, Zhengliang and Wu, Zihao and Li, Yiwei and Yang, Tianze and Shu, Peng and Xu, Shaochen and Dai, Haixing and Zhao, Lin and Mai, Gengchen and others},
  journal={arXiv preprint arXiv:2401.11641},
  year={2024}
}

@article{li2024large,
  title={Large Language Models for Manufacturing},
  author={Li, Yiwei and Zhao, Huaqin and Jiang, Hanqi and Pan, Yi and Liu, Zhengliang and Wu, Zihao and Shu, Peng and Tian, Jie and Yang, Tianze and Xu, Shaochen and others},
  journal={arXiv preprint arXiv:2410.21418},
  year={2024}
}

@article{ma2024iterative,
  title={An Iterative Optimizing Framework for Radiology Report Summarization with ChatGPT},
  author={Ma, Chong and Wu, Zihao and Wang, Jiaqi and Xu, Shaochen and Wei, Yaonai and Liu, Zhengliang and Zeng, Fang and Jiang, Xi and Guo, Lei and Cai, Xiaoyan and others},
  journal={IEEE Transactions on Artificial Intelligence},
  year={2024},
  publisher={IEEE}
}

@article{zhenyuan2024analyzing,
  title={Analyzing Nobel Prize Literature with Large Language Models},
  author={Zhenyuan, Yang and Zhengliang, Liu and Jing, Zhang and Cen, Lu and Jiaxin, Tai and Tianyang, Zhong and Yiwei, Li and Siyan, Zhao and Teng, Yao and Qing, Liu and others},
  journal={arXiv preprint arXiv:2410.18142},
  year={2024}
}

@article{wang2024legal,
  title={Legal Evalutions and Challenges of Large Language Models},
  author={Wang, Jiaqi and Zhao, Huan and Yang, Zhenyuan and Shu, Peng and Chen, Junhao and Sun, Haobo and Liang, Ruixi and Li, Shixin and Shi, Pengcheng and Ma, Longjun and others},
  journal={arXiv preprint arXiv:2411.10137},
  year={2024}
}

@article{gwerevende2023safeguarding,
  title={Safeguarding intangible cultural heritage: exploring the synergies in the transmission of Indigenous languages, dance and music practices in Southern Africa},
  author={Gwerevende, Solomon and Mthombeni, Zama M},
  journal={International Journal of Heritage Studies},
  volume={29},
  number={5},
  pages={398--412},
  year={2023},
  publisher={Taylor \& Francis}
}

@article{liu2023transformation,
  title={Transformation vs tradition: Artificial general intelligence (agi) for arts and humanities},
  author={Liu, Zhengliang and Li, Yiwei and Cao, Qian and Chen, Junwen and Yang, Tianze and Wu, Zihao and Hale, John and Gibbs, John and Rasheed, Khaled and Liu, Ninghao and others},
  journal={arXiv preprint arXiv:2310.19626},
  year={2023}
}

@article{achiam2023gpt,
  title={Gpt-4 technical report},
  author={Achiam, Josh and Adler, Steven and Agarwal, Sandhini and Ahmad, Lama and Akkaya, Ilge and Aleman, Florencia Leoni and Almeida, Diogo and Altenschmidt, Janko and Altman, Sam and Anadkat, Shyamal and others},
  journal={arXiv preprint arXiv:2303.08774},
  year={2023}
}

@article{devlin2018bert,
  title={Bert: Pre-training of deep bidirectional transformers for language understanding},
  author={Devlin, Jacob},
  journal={arXiv preprint arXiv:1810.04805},
  year={2018}
}

@article{radford2019language,
  title={Language models are unsupervised multitask learners},
  author={Radford, Alec and Wu, Jeffrey and Child, Rewon and Luan, David and Amodei, Dario and Sutskever, Ilya and others},
  journal={OpenAI blog},
  volume={1},
  number={8},
  pages={9},
  year={2019}
}

@article{brown2020language,
  title={Language models are few-shot learners},
  author={Brown, Tom B},
  journal={arXiv preprint arXiv:2005.14165},
  year={2020}
}

@article{cho2014learning,
  title={Learning phrase representations using RNN encoder-decoder for statistical machine translation},
  author={Cho, Kyunghyun},
  journal={arXiv preprint arXiv:1406.1078},
  year={2014}
}

@article{graves2012long,
  title={Long short-term memory},
  author={Graves, Alex and Graves, Alex},
  journal={Supervised sequence labelling with recurrent neural networks},
  pages={37--45},
  year={2012},
  publisher={Springer}
}

@article{dubey2024llama,
  title={The llama 3 herd of models},
  author={Dubey, Abhimanyu and Jauhri, Abhinav and Pandey, Abhinav and Kadian, Abhishek and Al-Dahle, Ahmad and Letman, Aiesha and Mathur, Akhil and Schelten, Alan and Yang, Amy and Fan, Angela and others},
  journal={arXiv preprint arXiv:2407.21783},
  year={2024}
}

@article{lee2023multimodality,
  title={Multimodality of ai for education: Towards artificial general intelligence},
  author={Lee, Gyeong-Geon and Shi, Lehong and Latif, Ehsan and Gao, Yizhu and Bewersdorf, Arne and Nyaaba, Matthew and Guo, Shuchen and Wu, Zihao and Liu, Zhengliang and Wang, Hui and others},
  journal={arXiv preprint arXiv:2312.06037},
  year={2023}
}

@inproceedings{finn2017model,
  title={Model-agnostic meta-learning for fast adaptation of deep networks},
  author={Finn, Chelsea and Abbeel, Pieter and Levine, Sergey},
  booktitle={International conference on machine learning},
  pages={1126--1135},
  year={2017},
  organization={PMLR}
}

@article{zhao2023brain,
  title={When brain-inspired ai meets agi},
  author={Zhao, Lin and Zhang, Lu and Wu, Zihao and Chen, Yuzhong and Dai, Haixing and Yu, Xiaowei and Liu, Zhengliang and Zhang, Tuo and Hu, Xintao and Jiang, Xi and others},
  journal={Meta-Radiology},
  pages={100005},
  year={2023},
  publisher={Elsevier}
}

@article{conneau2019unsupervised,
  title={Unsupervised cross-lingual representation learning at scale},
  author={Conneau, A},
  journal={arXiv preprint arXiv:1911.02116},
  year={2019}
}

@article{Liu2023,
  title={Transformation vs tradition: Artificial general intelligence (agi) for arts and humanities},
  author={Liu, Zhengliang and Li, Yiwei and Cao, Qian and Chen, Junwen and Yang, Tianze and Wu, Zihao and Hale, John and Gibbs, John and Rasheed, Khaled and Liu, Ninghao and others},
  journal={arXiv preprint arXiv:2310.19626},
  year={2023}
}

@article{popovic2021artificial,
  title={Artificial intelligence based writer identification generates new evidence for the unknown scribes of the Dead Sea Scrolls exemplified by the Great Isaiah Scroll (1QIsaa)},
  author={Popovi{\'c}, Mladen and Dhali, Maruf A and Schomaker, Lambert},
  journal={PloS one},
  volume={16},
  number={4},
  pages={e0249769},
  year={2021},
  publisher={Public Library of Science}
}

@article{assael2022restoring,
  title={Restoring and attributing ancient texts using deep neural networks},
  author={Assael, Yannis and Sommerschield, Thea and Shillingford, Brendan and Bordbar, Mahyar and Pavlopoulos, John and Chatzipanagiotou, Marita and Androutsopoulos, Ion and Prag, Jonathan and de Freitas, Nando},
  journal={Nature},
  volume={603},
  number={7900},
  pages={280--283},
  year={2022},
  publisher={Nature Publishing Group UK London}
}

@article{dhali2019binet,
  title={Binet: Degraded-manuscript binarization in diverse document textures and layouts using deep encoder-decoder networks},
  author={Dhali, Maruf A and de Wit, Jan Willem and Schomaker, Lambert},
  journal={arXiv preprint arXiv:1911.07930},
  year={2019}
}

@article{guan2024deciphering,
  title={Deciphering Oracle Bone Language with Diffusion Models},
  author={Guan, Haisu and Yang, Huanxin and Wang, Xinyu and Han, Shengwei and Liu, Yongge and Jin, Lianwen and Bai, Xiang and Liu, Yuliang},
  journal={arXiv preprint arXiv:2406.00684},
  year={2024}
}

@inproceedings{barucci2022ancient,
  title={Ancient Egyptian hieroglyphs segmentation and classification with convolutional neural networks},
  author={Barucci, Andrea and Canfailla, Chiara and Cucci, Costanza and Forasassi, Matteo and Franci, Massimiliano and Guarducci, Guido and Guidi, Tommaso and Loschiavo, Marco and Picollo, Marcello and Pini, Roberto and others},
  booktitle={International Conference Florence Heri-Tech: the Future of Heritage Science and Technologies},
  pages={126--139},
  year={2022},
  organization={Springer}
}

@article{koopmans2024performance,
  title={Performance Analysis of Handwritten Text Augmentation on Style-Based Dating of Historical Documents},
  author={Koopmans, Lisa and Dhali, Maruf A and Schomaker, Lambert},
  journal={SN Computer Science},
  volume={5},
  number={4},
  pages={397},
  year={2024},
  publisher={Springer}
}

@inproceedings{dhali2017digital,
  title={A digital palaeographic approach towards writer identification in the dead sea scrolls},
  author={Dhali, Maruf and He, Sheng and Popovic, Mladen and Tigchelaar, Eibert and Schomaker, Lambert},
  booktitle={International Conference on Pattern Recognition Applications and Methods 2017},
  pages={693--702},
  year={2017}
}

@book{saunt1999new,
  title={A new order of things: property, power, and the transformation of the Creek Indians, 1733-1816},
  author={Saunt, Claudio},
  year={1999},
  publisher={Cambridge University Press}
}

@incollection{saunt2017our,
  title={“Our Indians”: European Empires and the History of the Native American South},
  author={Saunt, Claudio},
  booktitle={The Atlantic in Global History},
  pages={63--79},
  year={2017},
  publisher={Routledge}
}

@book{saunt2005black,
  title={Black, White, and Indian: Race and the unmaking of an American family},
  author={Saunt, Claudio},
  year={2005},
  publisher={Oxford University Press, USA}
}

@book{saunt2020unworthy,
  title={Unworthy republic: The dispossession of Native Americans and the road to Indian Territory},
  author={Saunt, Claudio},
  year={2020},
  publisher={WW Norton \& Company}
}

@article{saunt2016age,
  title={The Age of Imperial Expansion, 1763--1821},
  author={Saunt, Claudio},
  journal={The Oxford Handbook of American Indian History},
  pages={77},
  year={2016},
  publisher={Oxford University Press}
}

@book{yang2021medicated,
  title={A Medicated Empire: The Pharmaceutical Industry and Modern Japan},
  author={Yang, Timothy M},
  year={2021},
  publisher={Cornell University Press}
}

@article{liu2023summary,
  title={Summary of chatgpt-related research and perspective towards the future of large language models},
  author={Liu, Yiheng and Han, Tianle and Ma, Siyuan and Zhang, Jiayue and Yang, Yuanyuan and Tian, Jiaming and He, Hao and Li, Antong and He, Mengshen and Liu, Zhengliang and others},
  journal={Meta-Radiology},
  pages={100017},
  year={2023},
  publisher={Elsevier}
}

@article{liu2024understanding,
  title={Understanding llms: A comprehensive overview from training to inference},
  author={Liu, Yiheng and He, Hao and Han, Tianle and Zhang, Xu and Liu, Mengyuan and Tian, Jiaming and Zhang, Yutong and Wang, Jiaqi and Gao, Xiaohui and Zhong, Tianyang and others},
  journal={arXiv preprint arXiv:2401.02038},
  year={2024}
}

@article{sommerschield2023machine,
  title={Machine learning for ancient languages: A survey},
  author={Sommerschield, Thea and Assael, Yannis and Pavlopoulos, John and Stefanak, Vanessa and Senior, Andrew and Dyer, Chris and Bodel, John and Prag, Jonathan and Androutsopoulos, Ion and de Freitas, Nando},
  journal={Computational Linguistics},
  volume={49},
  number={3},
  pages={703--747},
  year={2023},
  publisher={MIT Press One Broadway, 12th Floor, Cambridge, Massachusetts 02142, USA~…}
}

@inproceedings{Hedderich2020ASO,
  title={A Survey on Recent Approaches for Natural Language Processing in Low-Resource Scenarios},
  author={Michael A. Hedderich and Lukas Lange and Heike Adel and Jannik Strotgen and Dietrich Klakow},
  booktitle={North American Chapter of the Association for Computational Linguistics},
  year={2020},
  url={https://api.semanticscholar.org/CorpusID:225062337}
}

@article{LSSZ202408005,
author = {Yao, Dengfeng and Zhao, Yuan and Ye, Yurui and Rao, Gaoqi and ABULIZI Abudukelimu},
title = {Where Will Low-resource Languages Go in the Context of ChatGPT?},
journal = {Journal of Leshan Normal University},
volume = {39},
number = {08},
pages = {36-44},
year = {2024},
issn = {1009-8666},
doi = {10.16069/j.cnki.51-1610/g4.2024.08.005}
}

@article{ranathunga2023neural,
  title={Neural machine translation for low-resource languages: A survey},
  author={Ranathunga, Surangika and Lee, En-Shiun Annie and Prifti Skenduli, Marjana and Shekhar, Ravi and Alam, Mehreen and Kaur, Rishemjit},
  journal={ACM Computing Surveys},
  volume={55},
  number={11},
  pages={1--37},
  year={2023},
  publisher={ACM New York, NY}
}

@article{zoph2016transfer,
  title={Transfer learning for low-resource neural machine translation},
  author={Zoph, Barret and Yuret, Deniz and May, Jonathan and Knight, Kevin},
  journal={arXiv preprint arXiv:1604.02201},
  year={2016}
}

@article{gonen2020s,
  title={It's not Greek to mBERT: inducing word-level translations from multilingual BERT},
  author={Gonen, Hila and Ravfogel, Shauli and Elazar, Yanai and Goldberg, Yoav},
  journal={arXiv preprint arXiv:2010.08275},
  year={2020}
}

@ARTICLE{9392366,
  author={Zhang, Yu and Yang, Qiang},
  journal={IEEE Transactions on Knowledge and Data Engineering}, 
  title={A Survey on Multi-Task Learning}, 
  year={2022},
  volume={34},
  number={12},
  pages={5586-5609},
  doi={10.1109/TKDE.2021.3070203}}

@inproceedings{ragni2014data,
  title={Data augmentation for low resource languages},
  author={Ragni, Anton and Knill, Kate M and Rath, Shakti P and Gales, Mark JF},
  booktitle={INTERSPEECH 2014: 15th annual conference of the international speech communication association},
  pages={810--814},
  year={2014},
  organization={International Speech Communication Association (ISCA)}
}

@article{cekinel2024cross,
  title={Cross-Lingual Learning vs. Low-Resource Fine-Tuning: A Case Study with Fact-Checking in Turkish},
  author={Cekinel, Recep Firat and Karagoz, Pinar and Coltekin, Cagri},
  journal={arXiv preprint arXiv:2403.00411},
  year={2024}
}

@inproceedings{bimagambetova2023evaluating,
  title={Evaluating Large Language Models for Sentence Augmentation in Low-Resource Languages: A Case Study on Kazakh},
  author={Bimagambetova, Zhamilya and Rakhymzhanov, Dauren and Jaxylykova, Assel and Pak, Alexander},
  booktitle={2023 19th International Asian School-Seminar on Optimization Problems of Complex Systems (OPCS)},
  pages={14--18},
  year={2023},
  organization={IEEE}
}

@inproceedings{mani2023large,
  title={Large Language Models (LLMs): Representation Matters, Low-Resource Languages and Multi-Modal Architecture},
  author={Mani, Ganesh and Namomsa, Galane Basha},
  booktitle={2023 IEEE AFRICON},
  pages={1--6},
  year={2023},
  organization={IEEE}
}

@article{alam2024llms,
  title={LLMs for Low Resource Languages in Multilingual, Multimodal, and Dialectal Settings},
  author={Alam, Firoj and Chowdhury, Shammur Absar and Boughorbel, Sabri and Hasanain, Maram},
  journal={Proceedings of the 18th Conference of the European Chapter of the Association for Computational Linguistics},
  pages={27--33},
  year={2024}
}

@article{parida2024pretrain,
  title={Building Pre-train LLM Dataset for the Indic Languages: A Case Study on Hindi},
  author={Parida, Shantipriya and Panwar, Shakshi and Lata, Kusum and Mishra, Sanskruti and Sekhar, Sambit},
  journal={arXiv preprint arXiv:2407.09855},
  year={2024}
}

@ARTICLE{10288436,
  author={Zaikis, Dimitrios and Vlahavas, Ioannis},
  journal={IEEE Access}, 
  title={From Pre-Training to Meta-Learning: A Journey in Low-Resource-Language Representation Learning}, 
  year={2023},
  volume={11},
  number={},
  pages={115951-115967},
  doi={10.1109/ACCESS.2023.3326337}}

@misc{seo2024retrievalaugmenteddataaugmentationlowresource,
      title={Retrieval-Augmented Data Augmentation for Low-Resource Domain Tasks}, 
      author={Minju Seo and Jinheon Baek and James Thorne and Sung Ju Hwang},
      year={2024},
      eprint={2402.13482},
      archivePrefix={arXiv},
      primaryClass={cs.CL},
      url={https://arxiv.org/abs/2402.13482}, 
}

@misc{magueresse2020lowresourcelanguagesreviewpast,
      title={Low-resource Languages: A Review of Past Work and Future Challenges}, 
      author={Alexandre Magueresse and Vincent Carles and Evan Heetderks},
      year={2020},
      eprint={2006.07264},
      archivePrefix={arXiv},
      primaryClass={cs.CL},
      url={https://arxiv.org/abs/2006.07264}, 
}

@misc{terhoeve2022highresourcemethodologicalbiaslowresource,
      title={High-Resource Methodological Bias in Low-Resource Investigations}, 
      author={Maartje ter Hoeve and David Grangier and Natalie Schluter},
      year={2022},
      eprint={2211.07534},
      archivePrefix={arXiv},
      primaryClass={cs.CL},
      url={https://arxiv.org/abs/2211.07534}, 
}

@Article{languages9030082,
AUTHOR = {Lívio, Camila and Howe, Chad},
TITLE = {Text Mining Approaches to Language Use in Social Media: The Case of Portuguese Bué},
JOURNAL = {Languages},
VOLUME = {9},
YEAR = {2024},
NUMBER = {3},
ARTICLE-NUMBER = {82},
URL = {https://www.mdpi.com/2226-471X/9/3/82},
ISSN = {2226-471X},
DOI = {10.3390/languages9030082}
}

@phdthesis{romanenko2020robust,
  author       = {Aleksei Romanenko},
  title        = {Robust Speech Recognition for Low-Resource Languages},
  year         = {2020},
  school       = {Ulm University and ITMO University},
  address      = {Ulm, Germany and St. Petersburg, Russia},
  type         = {Doctoral Dissertation},
  note         = {Supervised by Prof. Dr.-Ing. Wolfgang Minker and Prof. Dr. Sc. Yuri N. Matveev},
  url          = {http://dx.doi.org/10.18725/OPARU-41801},
}

@INPROCEEDINGS{10485786,
  author={Eledath, Dhanva and Baby, Arun and Singh, Shatrughan},
  booktitle={2024 National Conference on Communications (NCC)}, 
  title={Robust Speech Recognition Using Meta-Learning for Low-Resource Accents}, 
  year={2024},
  volume={},
  number={},
  pages={1-6},
  doi={10.1109/NCC60321.2024.10485786}}

@article{nguyen2021survey,
  title={Survey of post-OCR processing approaches},
  author={Nguyen, Thi Tuyet Hai and Jatowt, Adam and Coustaty, Mickael and Doucet, Antoine},
  journal={ACM Computing Surveys (CSUR)},
  volume={54},
  number={6},
  pages={1--37},
  year={2021},
  publisher={ACM New York, NY, USA}
}

@article{memon2020handwritten,
  title={Handwritten optical character recognition (OCR): A comprehensive systematic literature review (SLR)},
  author={Memon, Jamshed and Sami, Maira and Khan, Rizwan Ahmed and Uddin, Mueen},
  journal={IEEE access},
  volume={8},
  pages={142642--142668},
  year={2020},
  publisher={IEEE}
}

@inproceedings{avyodri2022optical,
  title={Optical character recognition (ocr) for text recognition and its post-processing method: A literature review},
  author={Avyodri, Ridvy and Lukas, Samuel and Tjahyadi, Hendra},
  booktitle={2022 1st International Conference on Technology Innovation and Its Applications (ICTIIA)},
  pages={1--6},
  year={2022},
  organization={IEEE}
}

@inproceedings{wiechetek2024ethical,
  title={The Ethical Question--Use of Indigenous Corpora for Large Language Models},
  author={Wiechetek, Linda and Pirinen, Flammie A and Gaup, B{\o}rre and Trosterud, Trond and Kappfjell, Maja Lisa and Moshagen, Sjur},
  booktitle={Proceedings of the 2024 Joint International Conference on Computational Linguistics, Language Resources and Evaluation (LREC-COLING 2024)},
  pages={15922--15931},
  year={2024}
}

@article{tonja2024inkubalm,
  title={InkubaLM: A small language model for low-resource African languages},
  author={Tonja, Atnafu Lambebo and Dossou, Bonaventure FP and Ojo, Jessica and Rajab, Jenalea and Thior, Fadel and Wairagala, Eric Peter and Aremu, Anuoluwapo and Moiloa, Pelonomi and Abbott, Jade and Marivate, Vukosi and others},
  journal={arXiv preprint arXiv:2408.17024},
  year={2024}
}

@article{mahfuz2024too,
  title={Too late to train, too early to use? a study on necessity and viability of low-resource bengali llms},
  author={Mahfuz, Tamzeed and Dey, Satak Kumar and Naswan, Ruwad and Adil, Hasnaen and Sayeed, Khondker Salman and Shahgir, Haz Sameen},
  journal={arXiv preprint arXiv:2407.00416},
  year={2024}
}

@inproceedings{kholodna2024llms,
  title={Llms in the loop: Leveraging large language model annotations for active learning in low-resource languages},
  author={Kholodna, Nataliia and Julka, Sahib and Khodadadi, Mohammad and Gumus, Muhammed Nurullah and Granitzer, Michael},
  booktitle={Joint European Conference on Machine Learning and Knowledge Discovery in Databases},
  pages={397--412},
  year={2024},
  organization={Springer}
}

@article{wang2024improving,
  title={Improving Low-Resource Machine Translation Using Reinforcement Learning from Human Feedback.},
  author={Wang, Liqing and Xiao, Yiheng},
  journal={Intelligent Automation \& Soft Computing},
  volume={39},
  number={4},
  year={2024}
}

@article{hedderich2020survey,
  title={A survey on recent approaches for natural language processing in low-resource scenarios},
  author={Hedderich, Michael A and Lange, Lukas and Adel, Heike and Str{\"o}tgen, Jannik and Klakow, Dietrich},
  journal={arXiv preprint arXiv:2010.12309},
  year={2020}
}

@inproceedings{ogueji2021small,
  title={Small data? no problem! exploring the viability of pretrained multilingual language models for low-resourced languages},
  author={Ogueji, Kelechi and Zhu, Yuxin and Lin, Jimmy},
  booktitle={Proceedings of the 1st Workshop on Multilingual Representation Learning},
  pages={116--126},
  year={2021}
}

@article{hasan2024large,
  title={Do large language models speak all languages equally? a comparative study in low-resource settings},
  author={Hasan, Md Arid and Tarannum, Prerona and Dey, Krishno and Razzak, Imran and Naseem, Usman},
  journal={arXiv preprint arXiv:2408.02237},
  year={2024}
}

@inproceedings{dundjer2020automatic,
  title={Automatic machine translation of poetry and a low-resource language pair},
  author={Dun$\vec{d}$er, Ivan and Seljan, Sanja and Pavlovski, Marko},
  booktitle={2020 43rd International Convention on Information, Communication and Electronic Technology (MIPRO)},
  pages={1034--1039},
  year={2020},
  organization={IEEE}
}

@article{olsen1982meaning,
  title={The" Meaning" of a Literary Work},
  author={Olsen, Stein Haugom},
  journal={New Literary History},
  volume={14},
  number={1},
  pages={13--32},
  year={1982},
  publisher={JSTOR}
}

@article{karabayeva2024evaluating,
  title={Evaluating machine translation of literature through rhetorical analysis},
  author={Karabayeva, Irina and Kalizhanova, Anna},
  journal={Journal of Translation and Language Studies},
  volume={5},
  number={1},
  pages={1--9},
  year={2024}
}

@article{lin2020towards,
  title={Towards language service creation and customization for low-resource languages},
  author={Lin, Donghui and Murakami, Yohei and Ishida, Toru},
  journal={Information},
  volume={11},
  number={2},
  pages={67},
  year={2020},
  publisher={MDPI}
}

@article{huang2020cross,
  title={Cross-language transfer learning, continuous learning, and domain adaptation for end-to-end automatic speech recognition},
  author={Huang, Jocelyn and Kuchaiev, Oleksii and O'Neill, Patrick and Lavrukhin, Vitaly and Li, Jason and Flores, Adriana and Kucsko, Georg and Ginsburg, Boris},
  journal={arXiv preprint arXiv:2005.04290},
  year={2020}
}

@article{elewa2014features,
  title={Features of translating religious texts},
  author={Elewa, Abdelhamid},
  journal={Journal of translation},
  volume={10},
  number={1},
  pages={25--33},
  year={2014}
}

@article{jakel2002hypotheses,
  title={Hypotheses revisited: The cognitive theory of metaphor applied to religious texts},
  author={J{\"a}kel, Olaf},
  journal={Metaphorik. de},
  volume={2},
  number={1},
  pages={20--42},
  year={2002}
}

@article{agliz2015translation,
  title={Translation of Religious Texts: Difficulties and Challenges},
  author={Agliz, Rachid},
  journal={Arab World English Journal (AWEJ) Special Issue on Translation},
  year={2015}
}

@article{pasadika2007nirvana,
  title={Nirvana in Candrakirti's Prasannapada},
  author={Pasadika, Bhikkhu},
  journal={The Tibet Journal},
  volume={32},
  number={3},
  pages={64--68},
  year={2007},
  publisher={Library of Tibetan Works and Archives}
}

@article{brockington2004concept,
  title={The concept of" dharma" in the R{\=a}m{\=a}yaṇa},
  author={Brockington, John},
  journal={Journal of Indian Philosophy},
  volume={32},
  number={5/6},
  pages={655--670},
  year={2004},
  publisher={JSTOR}
}

@book{yancy2019buddhism,
  title={Buddhism and Whiteness: Critical Reflections},
  author={Yancy, G. and McRae, E. and Willis, J. and Suh, S. and Ann Gleig Ann Gleig, U. and Kalmanson, L. and Vesely-Flad, R.L. and Cassidy, L. and Medine, C.M.J. and Syedullah, J. and others},
  isbn={9781498581035},
  lccn={2019010958},
  series={Philosophy of Race},
  url={https://books.google.com/books?id=-e-YDwAAQBAJ},
  year={2019},
  publisher={Lexington Books}
}

@book{leblanc2012ancient,
  author       = {LeBlanc, John Randolph and Medine, Carolyn M. Jones},
  title        = {Ancient and Modern Religion and Politics: Negotiating Transitive Spaces and Hybrid Identities},
  publisher    = {Palgrave Macmillan},
  year         = {2012},
}

@article{jones_medine_2023,
  author       = {Jones Medine, Carolyn M.},
  title        = {T\&T Clark Handbook of African American Theology. Edited by Antonia Michelle Daymond, Frederick L. Ware, and Eric Lewis Williams. New York: Bloomsbury T\&T Clark, 2019. 464 pages. \$198.00.},
  journal      = {Horizons},
  volume       = {50},
  number       = {1},
  pages        = {230--232},
  year         = {2023},
  doi          = {10.1017/hor.2023.26}
}

@incollection{tehrani2023cultural,
    author = {Tehrani, Jamshid J.},
    isbn = {9780198869252},
    title = {The Cultural Transmission and Evolution of Folk Narratives},
    booktitle = {The Oxford Handbook of Cultural Evolution},
    publisher = {Oxford University Press},
    doi = {10.1093/oxfordhb/9780198869252.013.39},
    url = {https://doi.org/10.1093/oxfordhb/9780198869252.013.39},
    eprint = {https://academic.oup.com/book/0/chapter/411059290/chapter-ag-pdf/58712934/book\_45648\_section\_411059290.ag.pdf},
    year = {2023}
}

@article{li2024culturellm,
  title={Culturellm: Incorporating cultural differences into large language models},
  author={Li, Cheng and Chen, Mengzhou and Wang, Jindong and Sitaram, Sunayana and Xie, Xing},
  journal={arXiv preprint arXiv:2402.10946},
  year={2024}
}

@article{meaney2024evaluating,
  title={Evaluating and Adapting Large Language Models to Represent Folktales in Low-Resource Languages},
  author={Meaney, JA and Alex, Beatrice and Lamb, William},
  journal={arXiv preprint arXiv:2411.05593},
  year={2024}
}

@article{erasmo2020theatre,
  title={The Theatre of Pompey},
  author={Erasmo, Mario},
  journal={Memoirs of the American Academy in Rome},
  volume={65},
  pages={43--69},
  year={2020},
  publisher={JSTOR}
}

@article{rao2024normad,
  title={Normad: A benchmark for measuring the cultural adaptability of large language models},
  author={Rao, Abhinav and Yerukola, Akhila and Shah, Vishwa and Reinecke, Katharina and Sap, Maarten},
  journal={arXiv preprint arXiv:2404.12464},
  year={2024}
}

@article{DEANE1988325,
title = {Polysemy and cognition},
journal = {Lingua},
volume = {75},
number = {4},
pages = {325-361},
year = {1988},
issn = {0024-3841},
doi = {https://doi.org/10.1016/0024-3841(88)90009-5},
url = {https://www.sciencedirect.com/science/article/pii/0024384188900095},
author = {Paul D. Deane}
}

@phdthesis{Goel2023BeyondTS,
  title = {Beyond the Surface: A Computational Exploration of Linguistic Ambiguity},
  author = {Anmol Goel},
  year = {2023},
  school = {International Institute of Information Technology (IIIT), Hyderabad},
  type = {Master of Science Thesis},
  note = {Report No: IIIT/TH/2023/84},
  url = {https://api.semanticscholar.org/CorpusID:259371553}
}

@inproceedings{kibria-etal-2024-functional,
    title = "Detecting and Translating Language Ambiguity with Multilingual {LLM}s",
    author = "Mehrparvar, Behrang  and
      Pezzelle, Sandro",
    editor = {S{\"a}lev{\"a}, Jonne  and
      Owodunni, Abraham},
    booktitle = "Proceedings of the Fourth Workshop on Multilingual Representation Learning (MRL 2024)",
    month = nov,
    year = "2024",
    address = "Miami, Florida, USA",
    publisher = "Association for Computational Linguistics",
    url = "https://aclanthology.org/2024.mrl-1.26",
    pages = "310--323",
}

@article{Hutson2024,
  author       = {Hutson, James and Ellsworth, Pace and Ellsworth, Matt},
  title        = {Preserving Linguistic Diversity in the Digital Age: A Scalable Model for Cultural Heritage Continuity},
  year         = {2024},
  journal      = {Faculty Scholarship},
  volume       = {612},
  url          = {https://digitalcommons.lindenwood.edu/faculty-research-papers/612},
  note         = {Available online}
}

@misc{kirk2023personalisationboundsrisktaxonomy,
      title={Personalisation within bounds: A risk taxonomy and policy framework for the alignment of large language models with personalised feedback}, 
      author={Hannah Rose Kirk and Bertie Vidgen and Paul Röttger and Scott A. Hale},
      year={2023},
      eprint={2303.05453},
      archivePrefix={arXiv},
      primaryClass={cs.CL},
      url={https://arxiv.org/abs/2303.05453}, 
}

@misc{li2024cultureparkboostingcrossculturalunderstanding,
      title={CulturePark: Boosting Cross-cultural Understanding in Large Language Models}, 
      author={Cheng Li and Damien Teney and Linyi Yang and Qingsong Wen and Xing Xie and Jindong Wang},
      year={2024},
      eprint={2405.15145},
      archivePrefix={arXiv},
      primaryClass={cs.AI},
      url={https://arxiv.org/abs/2405.15145}, 
}

@misc{liu2023wereafraidlanguagemodels,
      title={We're Afraid Language Models Aren't Modeling Ambiguity}, 
      author={Alisa Liu and Zhaofeng Wu and Julian Michael and Alane Suhr and Peter West and Alexander Koller and Swabha Swayamdipta and Noah A. Smith and Yejin Choi},
      year={2023},
      eprint={2304.14399},
      archivePrefix={arXiv},
      primaryClass={cs.CL},
      url={https://arxiv.org/abs/2304.14399}, 
}

@misc{zhang2024goodllmsliterarytranslation,
      title={How Good Are LLMs for Literary Translation, Really? Literary Translation Evaluation with Humans and LLMs}, 
      author={Ran Zhang and Wei Zhao and Steffen Eger},
      year={2024},
      eprint={2410.18697},
      archivePrefix={arXiv},
      primaryClass={cs.CL},
      url={https://arxiv.org/abs/2410.18697}, 
}

@inproceedings{demidova-etal-2024-john,
    title = "John vs. Ahmed: Debate-Induced Bias in Multilingual {LLM}s",
    author = "Demidova, Anastasiia  and
      Atwany, Hanin  and
      Rabih, Nour  and
      Sha{'}ban, Sanad  and
      Abdul-Mageed, Muhammad",
    editor = "Habash, Nizar  and
      Bouamor, Houda  and
      Eskander, Ramy  and
      Tomeh, Nadi  and
      Abu Farha, Ibrahim  and
      Abdelali, Ahmed  and
      Touileb, Samia  and
      Hamed, Injy  and
      Onaizan, Yaser  and
      Alhafni, Bashar  and
      Antoun, Wissam  and
      Khalifa, Salam  and
      Haddad, Hatem  and
      Zitouni, Imed  and
      AlKhamissi, Badr  and
      Almatham, Rawan  and
      Mrini, Khalil",
    booktitle = "Proceedings of The Second Arabic Natural Language Processing Conference",
    month = aug,
    year = "2024",
    address = "Bangkok, Thailand",
    publisher = "Association for Computational Linguistics",
    url = "https://aclanthology.org/2024.arabicnlp-1.18",
    doi = "10.18653/v1/2024.arabicnlp-1.18",
    pages = "193--209",
}

@misc{jiang2024oraclesageunifiedvisuallinguisticunderstanding,
      title={OracleSage: Towards Unified Visual-Linguistic Understanding of Oracle Bone Scripts through Cross-Modal Knowledge Fusion}, 
      author={Hanqi Jiang and Yi Pan and Junhao Chen and Zhengliang Liu and Yifan Zhou and Peng Shu and Yiwei Li and Huaqin Zhao and Stephen Mihm and Lewis C Howe and Tianming Liu},
      year={2024},
      eprint={2411.17837},
      archivePrefix={arXiv},
      primaryClass={cs.CV},
      url={https://arxiv.org/abs/2411.17837}, 
}

@article{
doi:10.1073/pnas.2426815122,
author = {Gaurav Kamath  and Michelle Yang  and Siva Reddy  and Morgan Sonderegger  and Dallas Card },
title = {Semantic change in adults is not primarily a generational phenomenon},
journal = {Proceedings of the National Academy of Sciences},
volume = {122},
number = {31},
pages = {e2426815122},
year = {2025},
doi = {10.1073/pnas.2426815122},
URL = {https://www.pnas.org/doi/abs/10.1073/pnas.2426815122},
eprint = {https://www.pnas.org/doi/pdf/10.1073/pnas.2426815122}}

@misc{orife2020masakhanemachinetranslation,
      title={Masakhane -- Machine Translation For Africa}, 
      author={Iroro Orife and Julia Kreutzer and Blessing Sibanda and Daniel Whitenack and Kathleen Siminyu and Laura Martinus and Jamiil Toure Ali and Jade Abbott and Vukosi Marivate and Salomon Kabongo and Musie Meressa and Espoir Murhabazi and Orevaoghene Ahia and Elan van Biljon and Arshath Ramkilowan and Adewale Akinfaderin and Alp Öktem and Wole Akin and Ghollah Kioko and Kevin Degila and Herman Kamper and Bonaventure Dossou and Chris Emezue and Kelechi Ogueji and Abdallah Bashir},
      year={2020},
      eprint={2003.11529},
      archivePrefix={arXiv},
      primaryClass={cs.CL},
      url={https://arxiv.org/abs/2003.11529}, 
}

@misc{ginn2024glosslmmassivelymultilingualcorpus,
      title={GlossLM: A Massively Multilingual Corpus and Pretrained Model for Interlinear Glossed Text}, 
      author={Michael Ginn and Lindia Tjuatja and Taiqi He and Enora Rice and Graham Neubig and Alexis Palmer and Lori Levin},
      year={2024},
      eprint={2403.06399},
      archivePrefix={arXiv},
      primaryClass={cs.CL},
      url={https://arxiv.org/abs/2403.06399}, 
}

@inproceedings{calderon-etal-2025-alternative,
    title = "The Alternative Annotator Test for {LLM}-as-a-Judge: How to Statistically Justify Replacing Human Annotators with {LLM}s",
    author = "Calderon, Nitay  and
      Reichart, Roi  and
      Dror, Rotem",
    editor = "Che, Wanxiang  and
      Nabende, Joyce  and
      Shutova, Ekaterina  and
      Pilehvar, Mohammad Taher",
    booktitle = "Proceedings of the 63rd Annual Meeting of the Association for Computational Linguistics (Volume 1: Long Papers)",
    month = jul,
    year = "2025",
    address = "Vienna, Austria",
    publisher = "Association for Computational Linguistics",
    url = "https://aclanthology.org/2025.acl-long.782/",
    doi = "10.18653/v1/2025.acl-long.782",
    pages = "16051--16081",
    ISBN = "979-8-89176-251-0"
}

@article{artstein-poesio-2008-survey,
    title = "Survey Article: Inter-Coder Agreement for Computational Linguistics",
    author = "Artstein, Ron  and
      Poesio, Massimo",
    journal = "Computational Linguistics",
    volume = "34",
    number = "4",
    year = "2008",
    url = "https://aclanthology.org/J08-4004/",
    doi = "10.1162/coli.07-034-R2",
    pages = "555--596"
}

@inproceedings{Braylan_2022, series={WWW ’22},
   title={Measuring Annotator Agreement Generally across Complex Structured, Multi-object, and Free-text Annotation Tasks},
   url={http://dx.doi.org/10.1145/3485447.3512242},
   DOI={10.1145/3485447.3512242},
   booktitle={Proceedings of the ACM Web Conference 2022},
   publisher={ACM},
   author={Braylan, Alexander and Alonso, Omar and Lease, Matthew},
   year={2022},
   month=apr, pages={1720–1730},
   collection={WWW ’22} }

@inproceedings{hamilton-etal-2016-diachronic,
    title = "Diachronic Word Embeddings Reveal Statistical Laws of Semantic Change",
    author = "Hamilton, William L.  and
      Leskovec, Jure  and
      Jurafsky, Dan",
    editor = "Erk, Katrin  and
      Smith, Noah A.",
    booktitle = "Proceedings of the 54th Annual Meeting of the Association for Computational Linguistics (Volume 1: Long Papers)",
    month = aug,
    year = "2016",
    address = "Berlin, Germany",
    publisher = "Association for Computational Linguistics",
    url = "https://aclanthology.org/P16-1141/",
    doi = "10.18653/v1/P16-1141",
    pages = "1489--1501"
}

@inproceedings{schlechtweg-etal-2020-semeval,
    title = "{S}em{E}val-2020 Task 1: Unsupervised Lexical Semantic Change Detection",
    author = "Schlechtweg, Dominik  and
      McGillivray, Barbara  and
      Hengchen, Simon  and
      Dubossarsky, Haim  and
      Tahmasebi, Nina",
    editor = "Herbelot, Aurelie  and
      Zhu, Xiaodan  and
      Palmer, Alexis  and
      Schneider, Nathan  and
      May, Jonathan  and
      Shutova, Ekaterina",
    booktitle = "Proceedings of the Fourteenth Workshop on Semantic Evaluation",
    month = dec,
    year = "2020",
    address = "Barcelona (online)",
    publisher = "International Committee for Computational Linguistics",
    url = "https://aclanthology.org/2020.semeval-1.1/",
    doi = "10.18653/v1/2020.semeval-1.1",
    pages = "1--23"
}

@misc{amrami2019bettersubstitutionbasedwordsense,
      title={Towards better substitution-based word sense induction}, 
      author={Asaf Amrami and Yoav Goldberg},
      year={2019},
      eprint={1905.12598},
      archivePrefix={arXiv},
      primaryClass={cs.CL},
      url={https://arxiv.org/abs/1905.12598}, 
}

@inproceedings{Giulianelli_2020,
   title={Analysing Lexical Semantic Change with Contextualised Word Representations},
   url={http://dx.doi.org/10.18653/v1/2020.acl-main.365},
   DOI={10.18653/v1/2020.acl-main.365},
   booktitle={Proceedings of the 58th Annual Meeting of the Association for Computational Linguistics},
   publisher={Association for Computational Linguistics},
   author={Giulianelli, Mario and Del Tredici, Marco and Fernández, Raquel},
   year={2020} }

@misc{dicarlo2019trainingtemporalwordembeddings,
      title={Training Temporal Word Embeddings with a Compass}, 
      author={Valerio Di Carlo and Federico Bianchi and Matteo Palmonari},
      year={2019},
      eprint={1906.02376},
      archivePrefix={arXiv},
      primaryClass={cs.CL},
      url={https://arxiv.org/abs/1906.02376}, 
}

@article{kamath.2025.semantic,
author = {Gaurav Kamath  and Michelle Yang  and Siva Reddy  and Morgan Sonderegger  and Dallas Card },
title = {Semantic change in adults is not primarily a generational phenomenon},
journal = {Proceedings of the National Academy of Sciences},
volume = {122},
number = {31},
pages = {e2426815122},
year = {2025},
doi = {10.1073/pnas.2426815122},
URL = {https://www.pnas.org/doi/abs/10.1073/pnas.2426815122},
eprint = {https://www.pnas.org/doi/pdf/10.1073/pnas.2426815122},
}

@article{hofmann-etal-2024-geographic,
    title = "Geographic Adaptation of Pretrained Language Models",
    author = {Hofmann, Valentin  and
      Glava{\v{s}}, Goran  and
      Ljube{\v{s}}i{\'c}, Nikola  and
      Pierrehumbert, Janet B.  and
      Sch{\"u}tze, Hinrich},
    journal = "Transactions of the Association for Computational Linguistics",
    volume = "12",
    year = "2024",
    address = "Cambridge, MA",
    publisher = "MIT Press",
    url = "https://aclanthology.org/2024.tacl-1.23/",
    doi = "10.1162/tacl_a_00652",
    pages = "411--431"
}

@misc{wang2025geolocationawarerobustspokenlanguage,
      title={Geolocation-Aware Robust Spoken Language Identification}, 
      author={Qingzheng Wang and Hye-jin Shim and Jiancheng Sun and Shinji Watanabe},
      year={2025},
      eprint={2508.17148},
      archivePrefix={arXiv},
      primaryClass={cs.CL},
      url={https://arxiv.org/abs/2508.17148}, 
}

@inproceedings{vamvas-etal-2024-modular,
    title = "Modular Adaptation of Multilingual Encoders to Written {S}wiss {G}erman Dialect",
    author = {Vamvas, Jannis  and
      Aepli, No{\"e}mi  and
      Sennrich, Rico},
    editor = {V{\'a}zquez, Ra{\'u}l  and
      Mickus, Timothee  and
      Tiedemann, J{\"o}rg  and
      Vuli{\'c}, Ivan  and
      {\"U}st{\"u}n, Ahmet},
    booktitle = "Proceedings of the 1st Workshop on Modular and Open Multilingual NLP (MOOMIN 2024)",
    month = mar,
    year = "2024",
    address = "St Julians, Malta",
    publisher = "Association for Computational Linguistics",
    url = "https://aclanthology.org/2024.moomin-1.3/",
    pages = "16--23"
}

@inproceedings{khanuja-etal-2020-gluecos,
    title = "{GLUEC}o{S}: An Evaluation Benchmark for Code-Switched {NLP}",
    author = "Khanuja, Simran  and
      Dandapat, Sandipan  and
      Srinivasan, Anirudh  and
      Sitaram, Sunayana  and
      Choudhury, Monojit",
    editor = "Jurafsky, Dan  and
      Chai, Joyce  and
      Schluter, Natalie  and
      Tetreault, Joel",
    booktitle = "Proceedings of the 58th Annual Meeting of the Association for Computational Linguistics",
    month = jul,
    year = "2020",
    address = "Online",
    publisher = "Association for Computational Linguistics",
    url = "https://aclanthology.org/2020.acl-main.329/",
    doi = "10.18653/v1/2020.acl-main.329",
    pages = "3575--3585"
}

@inproceedings{mousi-etal-2025-aradice,
    title = "{A}ra{D}i{CE}: Benchmarks for Dialectal and Cultural Capabilities in {LLM}s",
    author = "Mousi, Basel  and
      Durrani, Nadir  and
      Ahmad, Fatema  and
      Hasan, Md. Arid  and
      Hasanain, Maram  and
      Kabbani, Tameem  and
      Dalvi, Fahim  and
      Chowdhury, Shammur Absar  and
      Alam, Firoj",
    editor = "Rambow, Owen  and
      Wanner, Leo  and
      Apidianaki, Marianna  and
      Al-Khalifa, Hend  and
      Eugenio, Barbara Di  and
      Schockaert, Steven",
    booktitle = "Proceedings of the 31st International Conference on Computational Linguistics",
    month = jan,
    year = "2025",
    address = "Abu Dhabi, UAE",
    publisher = "Association for Computational Linguistics",
    url = "https://aclanthology.org/2025.coling-main.283/",
    pages = "4186--4218"
}

@inproceedings{faisal-etal-24-dialectbenchy,
 title = {{DIALECTBENCH}: A {NLP} Benchmark for Dialects, Varieties, and Closely-Related Languages},
  author = {Faisal, Fahim and Ahia, Orevaoghene and Srivastava, Aarohi and Ahuja, Kabir and Chiang, David and Tsvetkov, Yulia and Anastasopoulos, Antonios},
  url={https://arxiv.org/pdf/2403.11009},
  year = {2024},
  booktitle = "Proceedings of ACL",
  address = "Bangkok, Thailand",
  publisher = "Association for Computational Linguistics",
  month = aug
}

@inproceedings{pires-etal-2019-multilingual,
  title = {How Multilingual is Multilingual {BERT}?},
  author = {Pires, Telmo and Schlinger, Eva and Garrette, Dan},
  booktitle = {Proceedings of the 57th Annual Meeting of the Association for Computational Linguistics},
  month = jul,
  year = {2019},
  address = {Florence, Italy},
  publisher = {Association for Computational Linguistics},
  url = {https://aclanthology.org/P19-1493/},
  doi = {10.18653/v1/P19-1493},
  pages = {4996--5001}
}

@inproceedings{foroutan-etal-2022-discovering,
  title = {Discovering Language-neutral Sub-networks in Multilingual Language Models},
  author = {Foroutan, Negar and Banaei, Mohammadreza and Lebret, R{\'e}mi and Bosselut, Antoine and Aberer, Karl},
  booktitle = {Proceedings of the 2022 Conference on Empirical Methods in Natural Language Processing},
  month = dec,
  year = {2022},
  address = {Abu Dhabi, United Arab Emirates},
  publisher = {Association for Computational Linguistics},
  url = {https://aclanthology.org/2022.emnlp-main.513/},
  doi = {10.18653/v1/2022.emnlp-main.513},
  pages = {7560--7575}
}

@inproceedings{tang-etal-2024-language,
  title = {Language-Specific Neurons: The Key to Multilingual Capabilities in Large Language Models},
  author = {Tang, Tianyi and Luo, Wenyang and Huang, Haoyang and Zhang, Dongdong and Wang, Xiaolei and Zhao, Xin and Wei, Furu and Wen, Ji-Rong},
  booktitle = {Proceedings of the 62nd Annual Meeting of the Association for Computational Linguistics (Volume 1: Long Papers)},
  month = aug,
  year = {2024},
  address = {Bangkok, Thailand},
  publisher = {Association for Computational Linguistics},
  url = {https://aclanthology.org/2024.acl-long.309/},
  doi = {10.18653/v1/2024.acl-long.309},
  pages = {5701--5715}
}

@inproceedings{pfeiffer-etal-2020-mad,
  title = {{MAD-X}: An Adapter-Based Framework for Multi-Task Cross-Lingual Transfer},
  author = {Pfeiffer, Jonas and Vuli{\'c}, Ivan and Gurevych, Iryna and Ruder, Sebastian},
  booktitle = {Proceedings of the 2020 Conference on Empirical Methods in Natural Language Processing (EMNLP)},
  month = nov,
  year = {2020},
  address = {Online},
  publisher = {Association for Computational Linguistics},
  url = {https://aclanthology.org/2020.emnlp-main.617/},
  doi = {10.18653/v1/2020.emnlp-main.617},
  pages = {7654--7673}
}

@article{shazeer2017moe,
  title = {Outrageously Large Neural Networks: The Sparsely-Gated Mixture-of-Experts Layer},
  author = {Shazeer, Noam and Mirhoseini, Azalia and Maziarz, Krzysztof and Davis, Andy and Le, Quoc and Hinton, Geoffrey and Dean, Jeff},
  journal = {arXiv preprint arXiv:1701.06538},
  year = {2017},
  url = {https://arxiv.org/abs/1701.06538}
}

@article{fedus2022switch,
  title = {Switch Transformers: Scaling to Trillion Parameter Models with Simple and Efficient Sparsity},
  author = {Fedus, William and Zoph, Barret and Shazeer, Noam},
  journal = {Journal of Machine Learning Research},
  volume = {23},
  number = {120},
  pages = {1--39},
  year = {2022},
  url = {https://jmlr.org/papers/v23/21-0998.html}
}

@article{arivazhagan2019massively,
  title = {Massively Multilingual Neural Machine Translation in the Wild: Findings and Challenges},
  author = {Arivazhagan, Naveen and Bapna, Ankur and Firat, Orhan and Lepikhin, Dmitry and Johnson, Melvin and Krikun, Maxim and Chen, Mia Xu and Cao, Yuan and Foster, George and Cherry, Colin and Macherey, Wolfgang and Chen, Zhifeng and Wu, Yonghui},
  journal = {arXiv preprint arXiv:1907.05019},
  year = {2019},
  url = {https://arxiv.org/abs/1907.05019}
}

@inproceedings{xhelili-etal-2024-breaking,
  title = {Breaking the Script Barrier in Multilingual Pre-Trained Language Models with Transliteration-Based Post-Training Alignment},
  author = {Xhelili, Orgest and Liu, Yihong and Schuetze, Hinrich},
  booktitle = {Findings of the Association for Computational Linguistics: EMNLP 2024},
  month = nov,
  year = {2024},
  address = {Miami, Florida, USA},
  publisher = {Association for Computational Linguistics},
  url = {https://aclanthology.org/2024.findings-emnlp.659/},
  doi = {10.18653/v1/2024.findings-emnlp.659},
  pages = {11283--11296}
}

@inproceedings{yu2020pcgrad,
  title = {Gradient Surgery for Multi-Task Learning},
  author = {Yu, Tianhe and Kumar, Saurabh and Gupta, Abhishek and Levine, Sergey and Hausman, Karol and Finn, Chelsea},
  booktitle = {Advances in Neural Information Processing Systems 33 (NeurIPS 2020)},
  year = {2020},
  url = {https://proceedings.neurips.cc/paper/2020/file/3fe78a8acf5fda99de95303940a2420c-Paper.pdf}
}

@inproceedings{winata-etal-2023-overcoming,
  title = {Overcoming Catastrophic Forgetting in Massively Multilingual Continual Learning},
  author = {Winata, Genta and Xie, Lingjue and Radhakrishnan, Karthik and Wu, Shijie and Jin, Xisen and Cheng, Pengxiang and Kulkarni, Mayank and Preotiuc-Pietro, Daniel},
  booktitle = {Findings of the Association for Computational Linguistics: ACL 2023},
  month = jul,
  year = {2023},
  address = {Toronto, Canada},
  publisher = {Association for Computational Linguistics},
  url = {https://aclanthology.org/2023.findings-acl.48/},
  doi = {10.18653/v1/2023.findings-acl.48},
  pages = {768--777}
}

@inproceedings{zhang-etal-2024-lightweight,
  title = {A Lightweight Mixture-of-Experts Neural Machine Translation Model with Stage-wise Training Strategy},
  author = {Zhang, Fan and Tu, Mei and Liu, Song and Yan, Jinyao},
  booktitle = {Findings of the Association for Computational Linguistics: NAACL 2024},
  month = jun,
  year = {2024},
  address = {Mexico City, Mexico},
  publisher = {Association for Computational Linguistics},
  url = {https://aclanthology.org/2024.findings-naacl.154/},
  doi = {10.18653/v1/2024.findings-naacl.154},
  pages = {2381--2392}
}

@article{chang2022geometry,
  title = {The Geometry of Multilingual Language Model Representations},
  author = {Chang, Tyler A. and Tu, Zhuowen and Bergen, Benjamin K.},
  journal = {arXiv preprint arXiv:2205.10964},
  year = {2022},
  doi = {10.48550/arXiv.2205.10964},
  url = {https://arxiv.org/abs/2205.10964}
}

\newpage

\appendix

\end{document}